\pdfoutput=1

\documentclass[11pt]{article}

\usepackage[final]{acl}
\usepackage{fancyhdr}
\fancypagestyle{firstpage}{
  \fancyhf{}
    \fancyhead[L]{\normalsize Accepted at EMNLP 2025}
     \fancyfoot[C]{\thepage}
}

\usepackage{times}
\usepackage{latexsym}
\usepackage{algorithm}
\usepackage{algorithmic}
\usepackage{multicol}
\usepackage{makecell}
\usepackage{multirow}
\usepackage{booktabs}
\usepackage{multirow}
\usepackage{amsmath}
\usepackage{amsfonts}
\usepackage{float}
\usepackage{subcaption}

\usepackage{placeins}
\usepackage{mathtools}

\usepackage{amsmath} 
\usepackage[T1]{fontenc}

\usepackage[utf8]{inputenc}

\usepackage{microtype}

\usepackage{inconsolata}

\usepackage{graphicx}
\usepackage{xcolor}
\definecolor{custompurple}{HTML}{9467bd} 
\definecolor{customgreen}{HTML}{2ca02c}  

\title{Aligning LLMs for Multilingual Consistency in Enterprise Applications}

\author{
 \textbf{Amit Agarwal\textsuperscript{}},
 \textbf{Hansa Meghwani\textsuperscript{}},
 \textbf{Hitesh Laxmichand Patel\textsuperscript{}}, \\
 \textbf{Tao Sheng\textsuperscript{}},
 \textbf{Sujith Ravi\textsuperscript{}},
 \textbf{Dan Roth\textsuperscript{}}
\\
\\
 \textsuperscript{}Oracle AI
\\
 \small{
   \textbf{Correspondence:} \href{mailto:amit.h.agarwal@oracle.com}{amit.h.agarwal@oracle.com}
 }
}

\begin{document}
\maketitle
\thispagestyle{firstpage}
\pagestyle{firstpage}
\begin{abstract}
Large language models (LLMs) remain unreliable for global enterprise applications due to substantial performance gaps between high-resource and mid/low-resource languages, driven by English-centric pretraining and internal reasoning biases. This inconsistency undermines customer experience and operational reliability in multilingual settings such as customer support, content moderation, and information retrieval. Even with advanced Retrieval-Augmented Generation (RAG) systems, we observe up to an 29\% accuracy drop in non-English languages compared to English.

We propose a practical, batch-wise alignment strategy for fine-tuning LLMs, leveraging semantically equivalent multilingual data in each training batch to directly align model outputs across languages. This approach improves non-English accuracy by up to 23.9\% without compromising English performance, model reasoning, or retrieval quality. Our method is simple to implement, scalable, and integrates seamlessly with existing LLM training \& deployment pipelines, enabling more robust and equitable multilingual AI solutions in industry. 
\end{abstract}

\section{Introduction}

The demand for multilingual natural language processing systems that perform reliably and consistently across diverse languages has grown rapidly in global industries such as customer support, content delivery networks, and information retrieval \cite{dua-etal-2025-speechweave,agarwal2024synthetic}. Despite advances in LLMs, significant performance disparities persist between high-resource languages (e.g., English) and low-resource languages (e.g., French, Arabic), primarily due to the pretraining data, 
favoring high-resource corpora \cite{xu2024survey}. This gap undermines user experience, restricts the accessibility of AI technologies, and limits operational effectiveness for organizations serving multilingual populations worldwide.

In practical deployments, LLMs often exhibit reduced accuracy and reasoning inconsistencies when handling non-English inputs, especially within RAG frameworks \cite{lewis2020retrieval, li2024bordirlines}. For example, a customer service chatbot may provide precise responses in English but fail to maintain equivalent quality in other languages, thereby eroding user trust and satisfaction. These challenges underscore an urgent need for AI systems that ensure equitable and consistent performance across languages.

Benchmarks like NoMIRACL \cite{thakur2023nomiracl} have advanced RAG across languages; however, these approaches typically optimize task-specific metrics without explicitly aligning the underlying multilingual reasoning and generation processes. Consequently, they fail to guarantee and evaluate consistency across languages. 
A key factor contributing to this problem is the English-centric internal reasoning of current LLMs~\cite{zhao2024large}, where LLMs internally “think” in English, even when processing inputs in other languages; exacerbating cross-lingual inconsistencies, reliability in multilingual applications. Existing approaches typically focus on external retrieval or translation to bridge this gap \cite{nie2022cross}.




To overcome these limitations, we propose a novel fine-tuning alignment strategy that explicitly aligns LLMs internal reasoning and generation processes across languages by leveraging semantically equivalent multilingual data within every training batch. Our approach reduces reliance on English-centric reasoning and improves cross-lingual consistency without introducing external translation or retrieval components. 
Our key contributions are summarized as follows:

\begin{itemize}
    \item We empirically characterize and quantify the performance gap in LLM reasoning and generation across languages for semantically identical content, highlighting its practical impact.
    \item We propose a simple yet effective batch-wise alignment strategy that fine-tunes LLMs to align internal multilingual generation processes without relying on external translation or retrieval systems.
    \item Our method achieves substantial improvements, up to \textbf{23.9\%}, in non-English task accuracy, narrowing the performance gap with English and enhancing model consistency.

\end{itemize}

\section{Related Work}

LLMs have made substantial progress, yet still exhibit significant performance disparities between high-resource and low-resource languages, in part due to English-centric pretraining and internal reasoning biases~\citep{zhao2024large, xu2024survey}. 

Recent research has further explored this space by penalizing inconsistent response languages~\citep{zhang2024respond}, applying cross-lingual instruction tuning with translation and mixed-task datasets~\citep{zhu2023extrapolating, lin2024crossin}, and leveraging preference optimization with translation and reward modeling~\citep{she2024mapo, dang2024rlhf}. In contrast, our approach achieves direct alignment of LLM reasoning and generation across languages by leveraging semantically equivalent multilingual data within each training batch, without relying on explicit translation, preference data, or reward models. This enables a simple and scalable alignment strategy suitable for industry ~\citep{zeng-etal-2025-converging}. See Appendix~\ref{sec:appendix_related} for an extended review.

\section{Method}


We now describe our batch-alignment approach, 
dataset construction, and the training paradigm. 

\subsection{Batch Alignment for Cross-Lingual Consistency}

\begin{figure}[!th]
  \centering
  \includegraphics[width=\columnwidth]{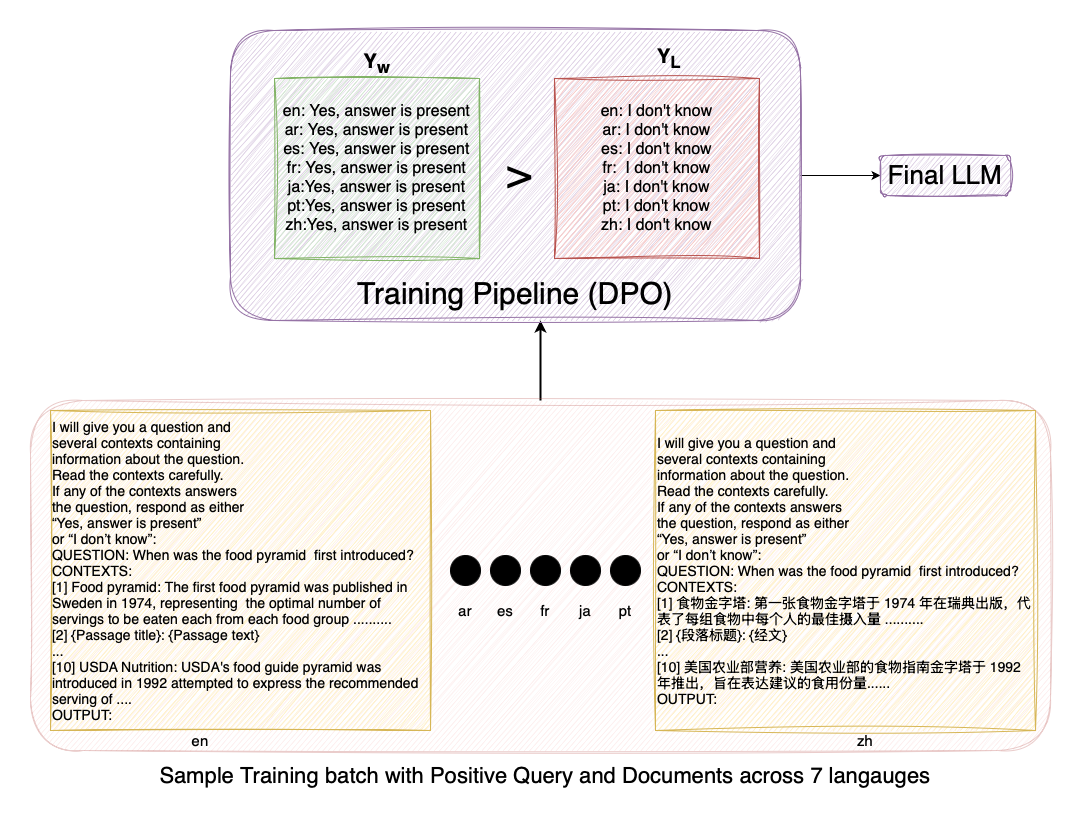}
  \caption{Highlights the training paradigm with the proposed batch-composition technique for a batch with Positive Samples. 
 The batch consists of training samples, with the same Query-Document composition across English and non-English to maintain same semantic consistency during training, where the LLM is expected to respond; "Yes, answer is present" or "I don't know" for the training samples.}
  \label{fig:training_pipeline}
  \vspace{-1.2em}
\end{figure}

The core of our approach is a \textit{batch alignment} strategy that modifies the 
batching process to explicitly encourage cross-lingual consistency. During training, each batch contains semantically equivalent instances of the same topic presented in different languages (Fig. \ref{fig:training_pipeline}). This setup exposes the model to the same semantic problem across multiple languages simultaneously, promoting aligned internal reasoning \& generation, reducing dependence on English-centric intermediate steps.

\paragraph{Batch Composition}  
Each training batch includes $k$ samples of the same topic, each in a different language. The batch size is determined by the number of languages included and controlled via ablation studies. This composition enforces the model to generate consistent outputs for semantically equivalent inputs across languages within a single update.

\subsection{Multilingual Dataset Construction} 


To enable systematic evaluation of multilingual reasoning and generation, we construct a multilingual question-answering dataset based on the NoMIRACL RAG framework. The dataset consists 500 documents, across diverse business-related topics spanning domains such as science, technology, advertising, and marketing. Each topic is translated by humans into six non-English languages (Arabic, Spanish, French, Japanese, Portuguese, and Chinese) in addition to English, ensuring semantic equivalence across languages.

For each topic, two distinct queries are designed that can be answered using a single relevant document. The dataset is constructed as follows:
\begin{itemize}
    \item \textbf{Positive Samples:} Each sample consists of a query along with one relevant document \& nine irrelevant documents selected from the same business domain but unrelated to the query, to simulate challenging retrieval conditions.
    
    \item \textbf{Negative Samples:} Each sample consists of a query along with ten irrelevant documents from the same domain, ensuring that no document directly answers the query.
\end{itemize}

\begin{figure}[!t]
  \centering
  \includegraphics[width=\columnwidth]{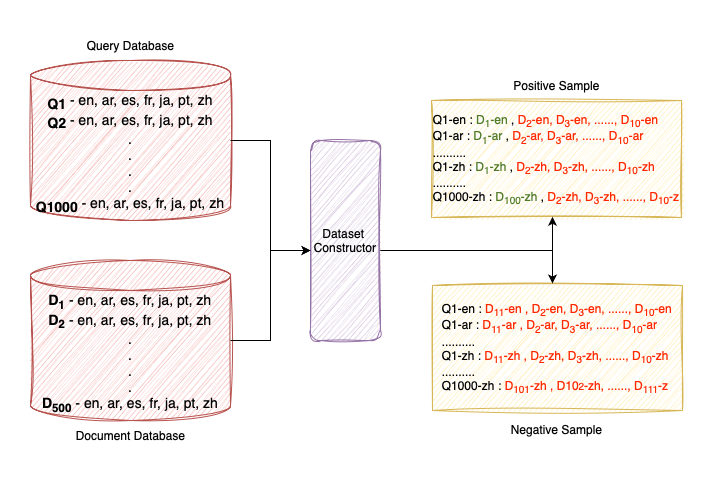}
  \caption{Highlights the high-level data-pipeline used to create the dataset. Each query is mapped by the Dataset Constructor to create a positive \& negative sample with corresponding documents of the respective language.}
  \label{fig:data_construction}
  \vspace{-1.5em}
\end{figure}

This design results in 2,000 samples (positive \& negative) per language. To ensure a fair cross-lingual evaluation, all contextual information, including queries and documents, is semantically consistent across languages. This controlled data set facilitates a rigorous analysis of multilingual reasoning and generation performance. Figure~\ref{fig:data_construction} illustrates our pipeline for constructing datasets, where queries and documents from human-annotated and translated databases are combined to form structured positive and negative samples for batch alignment training in a RAG setup. Additional details of prompt design and design rational are in the Appendix \ref{sec:appendix_prompt} \& \ref{sec:appendix_design}.

\subsection{Alignment Finetuning}

We adopt recent preference optimization methods: Direct Preference Optimization (DPO) \citep{rafailov2024direct} and Odds-Ratio Preference Optimization (ORPO) \citep{hong2024reference}, to optimize cross-lingual alignment. These approaches use pairwise preference signals derived from multilingual sample rankings within batches to guide the model towards consistent reasoning across languages. Inspired by \citet{shaham2024multilingual}, who demonstrated that minimal multilingual instruction tuning yields substantial cross-lingual generalization, we extend this principle to batch-level optimization, leveraging fine-grained multilingual supervision.

\vspace{-0.5em}

\paragraph{Training Procedure}  
Models are fine-tuned under the batch alignment paradigm with DPO and ORPO objectives. Standard finetuned models (uncoupled) use the same optimization methods, but with standard batch composition where samples from different topics and languages are randomly shuffled, providing no explicit multilingual alignment signal. Our method only changes batch composition; there are no extra critics/reward calls, no new parameters, and no inference overhead. Infrastructure details are available at Appendix \ref{sec:appendix_hardware}.
\vspace{-0.5em}
\section{Experimental Setup}

We describe our controlled RAG experimental setup, evaluation metrics, and the core research questions; see Appendix~\ref{sec:appendix_setup} for further details.

\subsection{Controlled RAG Setup}

To isolate the effect of our alignment strategy on multilingual generation, we conduct all evaluations within a RAG framework where the retrieval component is fixed and identical across languages. Specifically, for each query, the same set of documents is used, translated equivalently across languages, ensuring that the quality of the retrieval does not confound the performance of the generation. Furthermore, based on the documents in the context, the LLM has to respond according to the prompted instruction (see Appendix \ref{sec:appendix_prompt}). 

\paragraph{Instruction Language Control.} We fix the instruction/query template to English to control prompt-style variance and isolate cross-lingual conditioning from the content language. Documents are fully localized per language; only the instruction remains constant. This removes prompt-translation confounds in our controlled RAG probe — retrieval is fixed and identical across languages, so differences reflect generation/alignment rather than retriever noise or prompt phrasing. The sampler itself is anchor-agnostic, but including English as a high-resource anchor yields the best non-English gains (see Fig. \ref{fig:w_wo_english}), while English remains stable. Extended discussion and localized-prompt variants appear in Appendix \ref{sec:appendix_design}.



\paragraph{Evaluation Setup.}

Our dataset of 2,000 samples per language is split into 70\% training \& 30\% testing, balanced across positive and negative samples. Accuracy is used as the primary evaluation metric due to the binary classification nature of the task: the model must determine if the answer is present or absent based on the query-document pair.

\begin{figure*}[t]
  \centering
  \includegraphics[width=\textwidth]{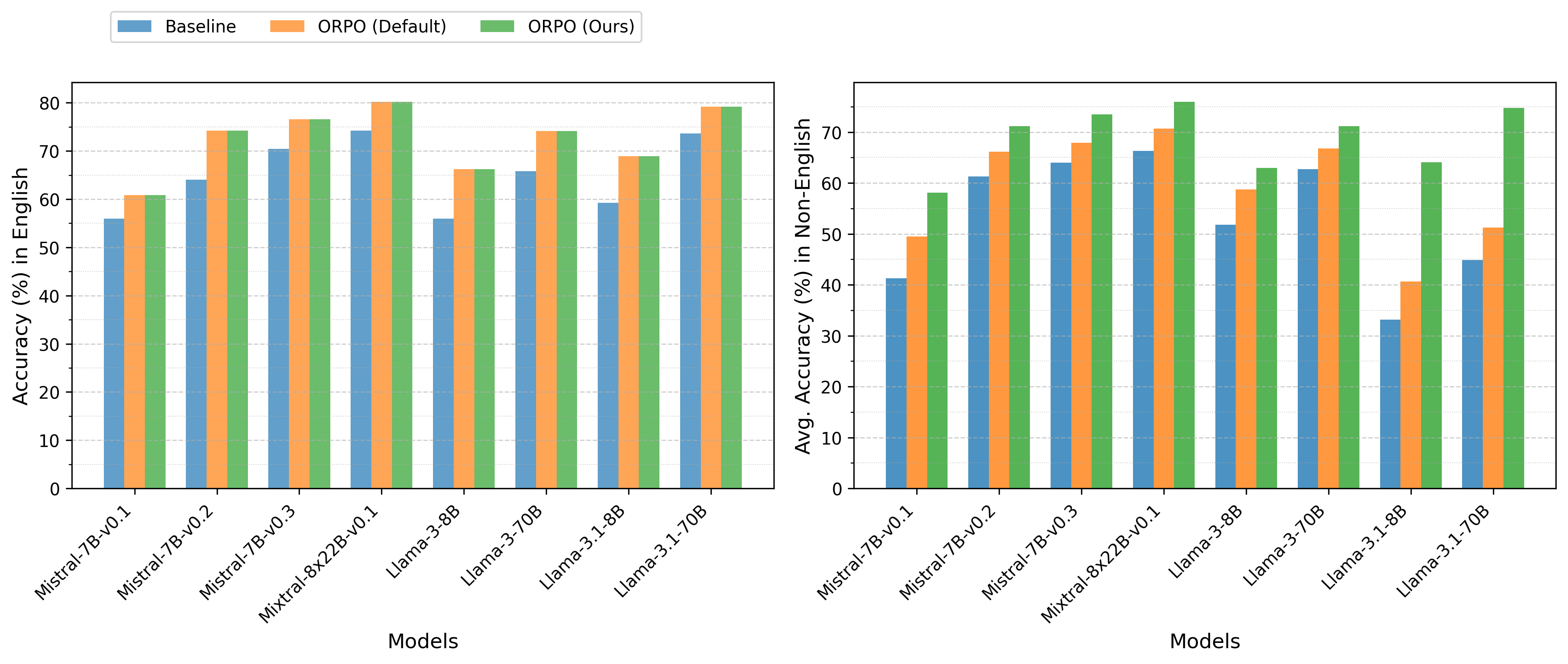}
  \caption{
LLM (instruct versions) accuracy in English (left) and averaged over six non-English languages (right) for Baseline, Default ORPO (uncoupled), and ORPO with batch alignment (Ours). Batch alignment substantially improves non-English accuracy and cross-lingual consistency. See Table~\ref{tab3} for full results.
}
  \label{fig:comparison_orpo}
\end{figure*}

\paragraph{Rationale for Accuracy Metric.}  
Given the controlled binary-response format (\ref{sec:appendix_prompt}) (i.e., “Yes, the answer is present” or “I don’t know”) uniformly applied across languages, exact match accuracy provides a direct, interpretable measure of reasoning and generation correctness. Since the retrieval is fixed, accuracy differences directly reflect the model’s multilingual reasoning and generation consistency. Thus, additional multilingual consistency metrics are unnecessary in this controlled setting.

\paragraph{Statistical Reporting.}  
We report point estimates (accuracy/exact match (EM) \%, perplexity) and compute 95\% bootstrap confidence intervals ($B{=}1000$) for all results. Pairwise comparisons on the same test items use McNemar’s test for accuracy/EM (significance at $p<0.01$) and paired bootstrap CIs for perplexity deltas. Claims of improvement are made only when the corresponding test is significant. \textit{Rest (avg.)} is the macro-average over non-English languages. CIs are displayed in Table [X]; due to space, other tables/figures report point estimates, computed with the same procedure.

\subsection{Research Questions}

To rigorously evaluate the effectiveness and generalizability of our batch alignment strategy, we organize our experiments around the following research questions:

\noindent\textbf{RQ1:} Does the proposed batch alignment method improve multilingual consistency in reasoning and generation across languages?  

\noindent\textbf{RQ2:} Can the alignment technique scale to unseen languages under the same task ?  

\noindent\textbf{RQ3:} Does the proposed alignment approach generalize to out-of-domain tasks beyond RAG?

\noindent\textbf{RQ4:} Does batch-alignment enhance other language modeling aspects such as fluency and semantic understanding?  


    





We 
conduct ablation studies to assess the influence of English as an anchor language, batch size, \& the use of machine-translated data for alignment. 

\vspace{-0.5em}
\section{Results}


We present empirical  results for each research question. Baseline LLMs exhibit substantial multilingual inconsistency in the RAG setup, with an 
accuracy drop up to 29\% in non-English languages (Table~\ref{tab1}, Appendix~\ref{sec:appendix_results}). The following subsections detail how our batch alignment strategy mitigates these gaps.



\subsection{RQ1: Impact of Batch Alignment on Multilingual Consistency}

We evaluated whether our proposed batch alignment strategy effectively improves multilingual consistency within RAG setups. Figure \ref{fig:comparison_orpo} compares the performance of three distinct setups: Baseline (pre-trained model), Default ORPO fine-tuning (uncoupled), and ORPO with batch alignment (ours): on English (left) and non-English languages (right).

\noindent \textbf{English Results:} ORPO fine-tuning significantly improves English accuracy over the baseline across all models. However, the incremental improvement from the default ORPO to ORPO with batch-alignment (ours)  is minimal, indicating limited additional benefit for high-resource languages that are already well represented during pre-training.

\noindent \textbf{Non-English Results:} In contrast, our batch alignment strategy yields substantial gains in multilingual accuracy compared to the default ORPO fine-tuning. Models such as Llama-3.1, which initially exhibit large multilingual performance gaps, benefit the most from our alignment method. This trend is consistent across various models, emphasizing the universal advantage of explicit multilingual alignment in reducing inherent model biases.
Table \ref{tab2} and Table \ref{tab3} in Appendix \ref{sec:appendix_results} provide detailed results across individual languages and models, further reinforcing these findings.



\subsection{RQ2: Generalization to Unseen Languages}
\label{sec:rq2}

We evaluate generalization on four unseen languages (Thai, Vietnamese, Hungarian, Romanian). ORPO with batch-alignment (ours) consistently outperform baselines in these languages (Table~\ref{tab:unseen_rag}); for example, Mixtral-8x22B and Llama-3-70B achieve accuracy gains of up to 4.1\%. This demonstrates that our batch alignment strategy enhances intrinsic cross-lingual generalization beyond the languages seen during training. Table~\ref{tab:unseen-langs-ci} in Appendix \ref{sec:appendix_results} expands the per language break-up and confidence-interval results.

\subsection{RQ3: Generalization to Out-of-Domain Tasks}

\begin{table}[!t]
\centering
\footnotesize
\scalebox{0.80}{\begin{tabular}{lcc|cc}
\toprule
\multirow{2}{*}{Model} & \multicolumn{2}{c|}{English} & \multicolumn{2}{c}{Rest (avg.)} \\
\cmidrule(lr){2-5}

& Baseline & ORPO (ours) & Baseline & ORPO (ours) \\
\midrule
Mixtral-8x22B & 74.2 & 80.2 & 41.7 & 44.3 \\
Llama-3-1-70B & 73.6 & 79.2 & 33.7 & 37.8 \\
Llama-3-70B   & 65.8 & 74.1 & 32.6 & 35.9 \\
\bottomrule
\end{tabular}}
\caption{Accuracy (\%) of baseline and ORPO with batch alignment (ours) on English and the average across four unseen languages (Rest (avg.)) demonstrating the cross-lingual generalization enabled by our alignment strategy.
}
\vspace{-1.5em}
\label{tab:unseen_rag}
\end{table}

Next, we validate our method's robustness by evaluating on more complex and diverse multilingual benchmarks such as Multilingual MMLU (MMMLU) and MGSM, which require extensive reasoning beyond retrieval-based QA. Table \ref{tab:mmlu} clearly shows that our aligned models consistently outperform the pre-trained baselines on the MMMLU benchmark. We observe average performance gains of 3.9\% for DPO (ours) and 4.2\% for ORPO (ours) across multiple models, highlighting our method's robust generalization and its effectiveness in enhancing complex multilingual reasoning.

Similarly, on the challenging MGSM benchmark (Table \ref{tab:mgsm}), our alignment strategy significantly boosts exact-match accuracy across all models. Larger models like Llama-3.1-70B achieve striking improvements (from 57.2\% to 74.9\%), underscoring the scalability and effectiveness of our batch alignment approach for sophisticated multilingual reasoning tasks.

\begin{table}[th!]
    \centering
    \scalebox{0.8}{\begin{tabular}{lcccc}
        \toprule
         \textbf{Models} &\textbf{Baseline} &\textbf{\makecell{DPO \\ (Ours)}} & \textbf{\makecell{ORPO \\ (Ours)}}  \\
        \midrule
      Mistral-7B-v0.1 & 45.6 & 48.8 & 48.8 \\
      
       Mistral-7B-v0.2 & 47.7 & 49.3 & 50.0 \\
     
         Mistral-7B-v0.3 & 50.7  & 52.6  & 53.1  \\
       
         Mixtral-8x22B-v0.1 & 75.2 & 77.2 & 77.6 \\
       
        Llama-3-8B & 49.0  & 59.4 & 59.5  \\
       
         Llama-3-70B & 73.7  & 76.5 & 76.8  \\
      
         Llama-3.1-8B & 61.8  & 68.4 & 68.4  \\
       
        Llama-3.1-70B & 79.9  & 82.6 & 83.0  \\
        \hline
    \end{tabular}}
    \caption{Accuracy of baseline \& batch-aligned models (Ours) on the MMMLU benchmark, highlighting consistent gains from our approach.}\label{tab:mmlu}
\vspace{-1.5em}
\end{table}

\begin{table}[!th]
\centering
\scalebox{0.85}{
\begin{tabular}{lcc}
\toprule
\textbf{Model} & \begin{tabular}[c]{@{}c@{}}Avg. Perf\\(Baseline)\end{tabular} & \begin{tabular}[c]{@{}c@{}}Avg. Perf\\(ORPO Ours)\end{tabular} \\
\midrule
Mistral-7B-v0.1    & 14.4 & 21.3 \\
Mistral-7B-v0.2    & 19.5 & 25.0 \\
Mistral-7B-v0.3    & 15.3 & 27.0 \\
Mixtral-8x22B-v0.1 & 26.7 & 33.8 \\
Llama-3-8B         & 28.1 & 34.9 \\
Llama-3-70B        & 67.9 & 71.8 \\
Llama-3.1-8B       & 30.1 & 39.3 \\
Llama-3.1-70B      & 57.2 & 74.9 \\
\bottomrule
\end{tabular}
}
\caption{Average Exact Match Accuracy of each model on MGSM benchmark, baseline and batch-aligned (ORPO ours) model.}
\label{tab:mgsm}
\vspace{-1.5em}
\end{table}

\subsection{RQ4: Broader Language Modeling Improvements}

Finally, we investigate whether batch alignment enhances broader aspects of language modeling beyond accuracy, focusing specifically on fluency and semantic comprehension. To this end, we evaluate on LAMBADA (multilingual), a close-style benchmark measuring contextual fluency and prediction, and PAWS-X, which tests cross-lingual paraphrase recognition and semantic understanding.

\begin{table*}[!th]
\centering
\scalebox{0.85}{
\begin{tabular}{l|cc|cc|cc}
\toprule
\textbf{Model} 
& \multicolumn{2}{c|}{\textbf{LAMBADA Perplexity} $\downarrow$}
& \multicolumn{2}{c|}{\textbf{LAMBADA Accuracy (\%)} $\uparrow$}
& \multicolumn{2}{c}{\textbf{PAWS-X Accuracy (\%)} $\uparrow$} \\
\cline{2-7}
& Baseline & ORPO (ours) & Baseline & ORPO (ours) & Baseline & ORPO (ours) \\
\hline
Mistral-7B-v0.1    & 54.9 & 49.8 & 30.6 & 37.3 & 61.8 & 63.9 \\
Mistral-7B-v0.2    & 37.7 & 33.1 & 34.0 & 38.5 & 64.3 & 69.9 \\
Mistral-7B-v0.3    & 29.2 & 24.8 & 36.2 & 39.9 & 63.2 & 69.7 \\
Mixtral-8x22B-v0.1 & 11.0 &  8.2 & 44.8 & 49.7 & 61.0 & 64.1 \\
Llama-3-8B         & 35.9 & 28.5 & 34.8 & 40.8 & 62.4 & 65.6 \\
Llama-3-70B        & 27.2 & 19.5 & 37.1 & 42.6 & 59.4 & 64.9 \\
Llama-3.1-8B       & 33.3 & 25.7 & 35.3 & 41.9 & 65.4 & 70.1 \\
Llama-3.1-70B      & 21.1 & 15.0 & 38.7 & 43.4 & 60.7 & 66.1 \\
\bottomrule
\end{tabular}
}
\caption{Performance comparison of Baseline and ORPO  batch-aligned (ours) models on two benchmarks: LAMBADA for Perplexity (lower is better) \& Accuracy (\%), and PAWS-X for Accuracy (\%).}
\label{tab:rq4}
\vspace{-1.5em}
\end{table*}

Table \ref{tab:rq4} reports performance on the LAMBADA (multilingual) and PAWS-X benchmarks. Our ORPO batch-aligned models consistently achieve lower perplexity scores and higher accuracy on LAMBADA, demonstrating improved fluency and coherence in multilingual text generation. Similarly, PAWS-X accuracy increases significantly across all tested models, indicating enhanced semantic understanding and better multilingual paraphrase recognition capabilities.



These broader language modeling improvements affirm the comprehensive benefits of our batch alignment method, extending beyond our controlled binary RAG probe: the same gains hold on \textsc{MMMLU}, \textsc{MGSM}, \textsc{PAWS-X}, and \textsc{LAMBADA} and are significant, enhancing the linguistic quality and semantic precision of multilingual models. We provide detailed results for each model and language in Appendix~\ref{sec:appendix_results}.

\subsection{Qualitative Error Analysis}
\label{sec:qual}
\noindent\textbf{Where errors arise.} In our controlled binary RAG probe, retrieval, index, and negatives are held fixed across languages, and outputs are constrained to \textit{Yes} / \textit{I don't know}. Residual mistakes therefore reflect \emph{generation-/alignment-/calibration}-side issues rather than retrieval or grading artifacts.

\noindent\textbf{Patterns relative to English.} Because the instruction/query template is fixed to English while documents are localized, we frequently observed \emph{co-occurrence of errors}: items wrong in English remain wrong in non-English runs, indicating model-intrinsic reasoning limits rather than language-specific lexical issues. Conversely, we do not observe items correct in non-English but wrong in English, consistent with English remaining a strong anchor.

\noindent\textbf{Data validation.} The documents/queries are factual and human-translated; we found no systematic cultural or language bias. We also verified the translated data against each model’s tokenizer and observed no unknown-token artifacts (e.g., \texttt{[unk]} token).

\vspace{-0.5em}
\section{Ablation Studies}

Key ablation findings are summarized below; full details are in Appendix~\ref{sec:appendix_ablation}.

\noindent\textbf{High-Resource Languages (\ref{sec:appendix_high_resource_language}):} 
Adding English as an anchor in training consistently boosts non-English accuracy, e.g., Mixtral-8x22B rises from 66.3\% to 76.0\%,
with an average gain of 9.8\% across models (Fig.~\ref{fig:w_wo_english}, Table \ref{tab6} \& \ref{tab7}).



\noindent\textbf{Machine-Translated Data (\ref{sec:appendix_machine_translated}):} Training with machine-translated data yields strong gains, though slightly below manual translations, confirming scalability and practicality for batch-alignment (Fig.~\ref{fig:machine_translated}).

\noindent\textbf{Batch Size (\ref{sec:appendix_batch_size}):} Larger batch sizes enhance multilingual accuracy by providing richer cross-lingual signals. Performance gains are consistent at higher batch sizes, underscoring the benefit of structured batch diversity in alignment (Fig.~\ref{fig:batch_size}).

\noindent \textbf{Model Size (\ref{sec:appendix_model_size}):}  We observe a weak positive correlation between model size and language improvement for both DPO and ORPO (Fig.~\ref{fig:corelation}), indicating that simply increasing model size does not fully close the multilingual performance gap.



\section{Discussion}



Our experiments demonstrate that batch-alignment is an effective and robust strategy for improving multilingual consistency in LLMs across varied tasks and benchmarks.


\noindent \textbf{Impact on Multilingual Consistency.}
Batch alignment consistently improves non-English accuracy without affecting English performance (RQ1). We find that including high-resource languages, especially English, as anchors in each batch is crucial for reducing multilingual disparities, likely by facilitating internal cross-lingual alignment within the model, For instance, including English increases non-English accuracy by an average of 9.8\% across models (see Fig.~\ref{fig:w_wo_english}).

\noindent \textbf{Generalization to Unseen Languages and Tasks.}
The ability to generalize to unseen languages (RQ2) further emphasizes the intrinsic multilingual capabilities enhanced by batch alignment. Our batch-aligned (ORPO) models not only outperform baseline models on seen languages but also exhibit measurable improvements on languages never encountered during fine-tuning. This outcome underscores the broader applicability of our technique beyond the original training context.

Moreover, results on out-of-domain benchmarks such as MMMLU \& MGSM (RQ3) provide strong evidence for the broader generalizability of our alignment approach. Our method notably enhances multilingual reasoning and knowledge-intensive tasks, showcasing its value beyond retrieval-based scenarios. Such generalization is particularly relevant for practical deployments where model adaptability across tasks and domains is crucial.

\noindent \textbf{Broader Implications for Language Modeling.}
Beyond accuracy improvements, batch alignment substantially improves model fluency, coherence, and semantic understanding, as demonstrated by reductions in perplexity on LAMBADA (multilingual) \& accuracy gains on the PAWS-X benchmark (RQ4). These broader improvements underscore that our alignment method fundamentally enhances multilingual language modeling rather than merely optimizing task-specific metrics. Thus, our technique contributes significantly towards building more reliable, coherent, and linguistically nuanced multilingual models.

\noindent \textbf{Insights from Ablation Studies.}
Our detailed ablation studies provide critical insights 
as we observe that larger batch sizes consistently enhance multilingual accuracy, with performance gains persisting across all tested batch sizes \& models, emphasizing the critical role of cross-lingual signal diversity within batches. 
Additionally, experiments with machine-translated data 
 indicate that while manual translation remains superior, machine-translated data also delivers substantial performance improvements (up to 15.1\%). This finding significantly enhances the practical feasibility of our method, especially in resource-limited scenarios.





\section{Conclusion}

We present a practical batch alignment strategy for fine-tuning large language models, substantially improving multilingual consistency in the generation component of RAG systems. Our approach, built on DPO and ORPO, yields up to a \textbf{23.9\% gain in non-English task accuracy} with no trade-off in English performance, closing the multilingual gap critical to real-world deployments.

This method is simple to integrate into existing LLM pipelines, requires only parallel or machine-translated data, and delivers immediate benefits for industry applications, enabling robust, equitable customer support, virtual assistants, and information retrieval in any language. Our ablation studies offer actionable guidance: English anchoring is essential, machine translation is viable for scaling, and batch size can be tuned to resource constraints.

By making multilingual alignment more accessible and practical, we empower global enterprises to serve diverse linguistic markets, reduce operational risk, and increase user satisfaction. 
Our findings suggest that LLMs possess latent multilingual reasoning and generation capabilities, which can be surfaced and enhanced through targeted alignment, opening new directions for robust, equitable AI across languages.
\section{Limitations}
While our findings demonstrate robust improvements, several limitations offer directions for future research. First, our study primarily focuses on high-resource languages as alignment anchors, raising the question of how low-resource language alignment might benefit from leveraging intermediate-resource languages. Second, despite notable improvements, the reliance on preference-based optimization methods such as DPO and ORPO introduces complexity in training setups, warranting exploration of simpler and more scalable optimization techniques. Finally, evaluating alignment strategies in highly noisy or real-world user-generated or synthetic multilingual datasets remains an open challenge, essential for practical deployments.

Furthermore, the study does not explore the impact of long, noisy sequences that are often present in real-world multilingual data. Such sequences may adversely affect the alignment process and model performance. Future work should address these gaps by expanding language coverage, evaluating the effect of noisy sequences, and assessing the viability of synthetic data for training.

\section{Ethical Considerations}
Our work aims to advance equitable access to AI by reducing performance disparities between high-resource and low-resource languages, thereby supporting fairer and more inclusive multilingual technologies. By promoting consistency across languages, our approach can help mitigate language-based biases that have historically limited access to high-quality AI services for underrepresented linguistic communities.

However, ethical deployment also requires careful attention to data quality \& representation. Overreliance on machine-translated/synthetic data could inadvertently amplify errors or encode unintended biases from dominant language corpora. We recommend ongoing benchmarking \& transparent reporting of multilingual model behavior, especially for languages with limited resources, to minimize potential harms \& maximize societal benefit.
\vspace{-1em}

\bibliography{acl}

\begin{thebibliography}{78}
\providecommand{\natexlab}[1]{#1}

\bibitem[{Agarwal(2021)}]{agarwal2021evaluate}
Amit Agarwal. 2021.
\newblock \href {https://doi.org/10.13140/RG.2.2.33887.53928} {Evaluate generalisation \& robustness of visual features from images to video}.
\newblock \emph{ResearchGate}.
\newblock Available at \url{https://doi.org/10.13140/RG.2.2.33887.53928}.

\bibitem[{Agarwal et~al.(2025{\natexlab{a}})Agarwal, Panda, Charles, Patel, Kumar, Pattnayak, Rafi, Kumar, Meghwani, Gupta, and Chae}]{agarwal-etal-2025-mvtamperbench}
Amit Agarwal, Srikant Panda, Angeline Charles, Hitesh~Laxmichand Patel, Bhargava Kumar, Priyaranjan Pattnayak, Taki~Hasan Rafi, Tejaswini Kumar, Hansa Meghwani, Karan Gupta, and Dong-Kyu Chae. 2025{\natexlab{a}}.
\newblock \href {https://doi.org/10.18653/v1/2025.findings-acl.963} {{MVT}amper{B}ench: Evaluating robustness of vision-language models}.
\newblock In \emph{Findings of the Association for Computational Linguistics: ACL 2025}, pages 18804--18828, Vienna, Austria. Association for Computational Linguistics.

\bibitem[{Agarwal et~al.(2024)Agarwal, Panda, and Pachauri}]{agarwal2024synthetic}
Amit Agarwal, Srikant Panda, and Kulbhushan Pachauri. 2024.
\newblock Synthetic document generation pipeline for training artificial intelligence models.
\newblock US Patent App. 17/994,712.

\bibitem[{Agarwal et~al.(2025{\natexlab{b}})Agarwal, Panda, and Pachauri}]{agarwal-etal-2025-fs}
Amit Agarwal, Srikant Panda, and Kulbhushan Pachauri. 2025{\natexlab{b}}.
\newblock \href {https://aclanthology.org/2025.coling-industry.9/} {{FS}-{DAG}: Few shot domain adapting graph networks for visually rich document understanding}.
\newblock In \emph{Proceedings of the 31st International Conference on Computational Linguistics: Industry Track}, pages 100--114, Abu Dhabi, UAE. Association for Computational Linguistics.

\bibitem[{Agarwal et~al.(2025{\natexlab{c}})Agarwal, Patel, Panda, Meghwani, Singh, Dua, Li, Sheng, Ravi, and Roth}]{agarwal2025rciscoreevaluatingglobal}
Amit Agarwal, Hitesh~Laxmichand Patel, Srikant Panda, Hansa Meghwani, Jyotika Singh, Karan Dua, Paul Li, Tao Sheng, Sujith Ravi, and Dan Roth. 2025{\natexlab{c}}.
\newblock \href {https://arxiv.org/abs/2509.23673} {Rci: A score for evaluating global and local reasoning in multimodal benchmarks}.
\newblock \emph{Preprint}, arXiv:2509.23673.

\bibitem[{Asai et~al.(2024)Asai, Kudugunta, Yu et~al.}]{asai2024buffet}
Akari Asai, Sneha Kudugunta, Xinyan~Velocity Yu, et~al. 2024.
\newblock Buffet: Benchmarking large language models for few-shot cross-lingual transfer.
\newblock \emph{Proceedings of NAACL 2024}, pages 1771--1800.

\bibitem[{Bai et~al.(2022)Bai, Jones, Ndousse, Askell, Chen, DasSarma, Drain, Fort, Ganguli, Henighan et~al.}]{bai2022training}
Yuntao Bai, Andy Jones, Kamal Ndousse, Amanda Askell, Anna Chen, Nova DasSarma, Dawn Drain, Stanislav Fort, Deep Ganguli, Tom Henighan, et~al. 2022.
\newblock Training a helpful and harmless assistant with reinforcement learning from human feedback.
\newblock \emph{arXiv preprint arXiv:2204.05862}.

\bibitem[{Cahyawijaya et~al.(2025)Cahyawijaya, Lovenia, Moniz, Wong, Farhansyah, Maung, Hudi, Anugraha, Habibi, Qorib, Agarwal, Imperial, Patel, Feliren, Nasution, Rufino, Winata, Rajagede, Catalan, Imam, Pattnayak, Pranida, Pratama, Bangera, Na-Thalang, Monderin, Song, Simon, Ng, Sapan, Rafi, Wang, Supryadi, Veerakanjana, Ittichaiwong, Roque, Vincentio, Kreangphet, Artkaew, Palgunadi, Yu, Hastuti, Nixon, Bangera, Lim, Khine, Zhafran, Ferdinan, Izzani, Singh, Evan, Krito, Anugraha, Ilasariya, Li, Daniswara, Tjiaranata, Yulianrifat, Udomcharoenchaikit, Ansori, Ihsani, Nguyen, Barik, Velasco, Genadi, Saha, Wei, Flores, Han, Santos, Lim, Phyo, Santos, Dwiastuti, Luo, Cruz, Hee, Hanif, Hakim, Sya{'}ban, Kerdthaisong, Miranda, Koto, Fatyanosa, Aji, Rosal, Kevin, Wijaya, Kampman, Zhang, Karlsson, and Limkonchotiwat}]{cahyawijaya-etal-2025-crowdsource}
Samuel Cahyawijaya, Holy Lovenia, Joel Ruben~Antony Moniz, Tack~Hwa Wong, Mohammad~Rifqi Farhansyah, Thant~Thiri Maung, Frederikus Hudi, David Anugraha, Muhammad Ravi~Shulthan Habibi, Muhammad~Reza Qorib, Amit Agarwal, Joseph~Marvin Imperial, Hitesh~Laxmichand Patel, Vicky Feliren, Bahrul~Ilmi Nasution, Manuel~Antonio Rufino, Genta~Indra Winata, Rian~Adam Rajagede, Carlos~Rafael Catalan, Mohamed Fazli~Mohamed Imam, Priyaranjan Pattnayak, Salsabila~Zahirah Pranida, Kevin Pratama, Yeshil Bangera, Adisai Na-Thalang, Patricia~Nicole Monderin, Yueqi Song, Christian Simon, Lynnette Hui~Xian Ng, Richardy~Lobo Sapan, Taki~Hasan Rafi, Bin Wang, Supryadi, Kanyakorn Veerakanjana, Piyalitt Ittichaiwong, Matthew~Theodore Roque, Karissa Vincentio, Takdanai Kreangphet, Phakphum Artkaew, Kadek~Hendrawan Palgunadi, Yanzhi Yu, Rochana~Prih Hastuti, William Nixon, Mithil Bangera, Adrian Xuan~Wei Lim, Aye~Hninn Khine, Hanif~Muhammad Zhafran, Teddy Ferdinan, Audra~Aurora Izzani, Ayushman Singh, Evan Evan, Jauza~Akbar Krito,
  Michael Anugraha, Fenal~Ashokbhai Ilasariya, Haochen Li, John~Amadeo Daniswara, Filbert~Aurelian Tjiaranata, Eryawan~Presma Yulianrifat, Can Udomcharoenchaikit, Fadil~Risdian Ansori, Mahardika~Krisna Ihsani, Giang Nguyen, Anab~Maulana Barik, Dan~John Velasco, Rifo~Ahmad Genadi, Saptarshi Saha, Chengwei Wei, Isaiah Edri~W. Flores, Kenneth Chen~Ko Han, Anjela Gail~D. Santos, Wan~Shen Lim, Kaung~Si Phyo, Tim Santos, Meisyarah Dwiastuti, Jiayun Luo, Jan Christian~Blaise Cruz, Ming~Shan Hee, Ikhlasul~Akmal Hanif, M.Alif~Al Hakim, Muhammad~Rizky Sya{'}ban, Kun Kerdthaisong, Lester James~Validad Miranda, Fajri Koto, Tirana~Noor Fatyanosa, Alham~Fikri Aji, Jostin~Jerico Rosal, Jun Kevin, Robert Wijaya, Onno~P. Kampman, Ruochen Zhang, B{\"o}rje~F. Karlsson, and Peerat Limkonchotiwat. 2025.
\newblock \href {https://doi.org/10.18653/v1/2025.acl-long.916} {Crowdsource, crawl, or generate? creating {SEA}-{VL}, a multicultural vision-language dataset for {S}outheast {A}sia}.
\newblock In \emph{Proceedings of the 63rd Annual Meeting of the Association for Computational Linguistics (Volume 1: Long Papers)}, pages 18685--18717, Vienna, Austria. Association for Computational Linguistics.

\bibitem[{Cobbe et~al.(2021)Cobbe, Kosaraju, Bavarian, Hilton, Nakano, Hesse, and Schulman}]{cobbe2021training}
Karl Cobbe, Vineet Kosaraju, Mohammad Bavarian, Jacob Hilton, Reiichiro Nakano, Christopher Hesse, and John Schulman. 2021.
\newblock \href {https://arxiv.org/abs/2110.14168} {Training verifiers to solve math word problems}.
\newblock \emph{Preprint}, arXiv:2110.14168.

\bibitem[{Conneau(2019)}]{conneau2019unsupervised}
A~Conneau. 2019.
\newblock Unsupervised cross-lingual representation learning at scale.
\newblock \emph{arXiv preprint arXiv:1911.02116}.

\bibitem[{Conneau et~al.(2018)Conneau, Lample, Rinott, Williams, Bowman, Schwenk, and Stoyanov}]{conneau2018xnli}
Alexis Conneau, Guillaume Lample, Ruty Rinott, Adina Williams, Samuel~R Bowman, Holger Schwenk, and Veselin Stoyanov. 2018.
\newblock Xnli: Evaluating cross-lingual sentence representations.
\newblock In \emph{Proceedings of the 2018 Conference on Empirical Methods in Natural Language Processing (EMNLP)}.

\bibitem[{Dang et~al.(2024)Dang, Ahmadian, Marchisio, Kreutzer, {\"U}st{\"u}n, and Hooker}]{dang2024rlhf}
John Dang, Arash Ahmadian, Kelly Marchisio, Julia Kreutzer, Ahmet {\"U}st{\"u}n, and Sara Hooker. 2024.
\newblock Rlhf can speak many languages: Unlocking multilingual preference optimization for llms, 2024.
\newblock \emph{URL https://arxiv. org/abs/2407.02552}.

\bibitem[{Devlin(2018)}]{devlin2018bert}
Jacob Devlin. 2018.
\newblock Bert: Pre-training of deep bidirectional transformers for language understanding.
\newblock \emph{arXiv preprint arXiv:1810.04805}.

\bibitem[{Dua et~al.(2025{\natexlab{a}})Dua, Mittal, Gupta, and Patel}]{dua-etal-2025-speechweave}
Karan Dua, Puneet Mittal, Ranjeet Gupta, and Hitesh~Laxmichand Patel. 2025{\natexlab{a}}.
\newblock \href {https://doi.org/10.18653/v1/2025.acl-industry.51} {{S}peech{W}eave: Diverse multilingual synthetic text {\&} audio data generation pipeline for training text to speech models}.
\newblock In \emph{Proceedings of the 63rd Annual Meeting of the Association for Computational Linguistics (Volume 6: Industry Track)}, pages 718--737, Vienna, Austria. Association for Computational Linguistics.

\bibitem[{Dua et~al.(2025{\natexlab{b}})Dua, Patel, Mittal, Gupta, Agarwal, Pabolu, Panda, Meghwani, Horwood, and Shah}]{dua2025flexdocparameterizedsamplingdiverse}
Karan Dua, Hitesh~Laxmichand Patel, Puneet Mittal, Ranjeet Gupta, Amit Agarwal, Praneet Pabolu, Srikant Panda, Hansa Meghwani, Graham Horwood, and Fahad Shah. 2025{\natexlab{b}}.
\newblock \href {https://arxiv.org/abs/2510.02133} {Flexdoc: Parameterized sampling for diverse multilingual synthetic documents for training document understanding models}.
\newblock \emph{Preprint}, arXiv:2510.02133.

\bibitem[{Dubey et~al.(2024)Dubey, Jauhri, Pandey, Kadian, Al-Dahle, Letman, Mathur, Schelten, Yang, Fan et~al.}]{dubey2024llama}
Abhimanyu Dubey, Abhinav Jauhri, Abhinav Pandey, Abhishek Kadian, Ahmad Al-Dahle, Aiesha Letman, Akhil Mathur, Alan Schelten, Amy Yang, Angela Fan, et~al. 2024.
\newblock The llama 3 herd of models.
\newblock \emph{arXiv preprint arXiv:2407.21783}.

\bibitem[{EleutherAI(2024)}]{eleutherai2024lambadamultilingual}
EleutherAI. 2024.
\newblock Lambada multilingual dataset.
\newblock \url{https://github.com/EleutherAI/lm-evaluation-harness/tree/main/lm_eval/tasks/lambada_multilingual}.
\newblock Accessed: 2025-06-07.

\bibitem[{Feng et~al.(2020)Feng, Yang, Cer, Arivazhagan, and Wang}]{feng2020language}
Fangxiaoyu Feng, Yinfei Yang, Daniel Cer, Naveen Arivazhagan, and Wei Wang. 2020.
\newblock Language-agnostic bert sentence embedding.
\newblock \emph{arXiv preprint arXiv:2007.01852}.

\bibitem[{Hong et~al.(2024{\natexlab{a}})Hong, Lee, and Thorne}]{hong2024orpo}
Jiwoo Hong, Noah Lee, and James Thorne. 2024{\natexlab{a}}.
\newblock Orpo: Monolithic preference optimization without reference model.
\newblock In \emph{Proceedings of the 2024 Conference on Empirical Methods in Natural Language Processing}, pages 11170--11189.

\bibitem[{Hong et~al.(2024{\natexlab{b}})Hong, Lee, and Thorne}]{hong2024reference}
Jiwoo Hong, Noah Lee, and James Thorne. 2024{\natexlab{b}}.
\newblock Reference-free monolithic preference optimization with odds ratio.
\newblock \emph{arXiv preprint arXiv:2403.07691}.

\bibitem[{Houlsby et~al.(2019)Houlsby, Giurgiu, Jastrzebski, Morrone, De~Laroussilhe, Gesmundo, Attariyan, and Gelly}]{houlsby2019parameter}
Neil Houlsby, Andrei Giurgiu, Stanislaw Jastrzebski, Bruna Morrone, Quentin De~Laroussilhe, Andrea Gesmundo, Mona Attariyan, and Sylvain Gelly. 2019.
\newblock Parameter-efficient transfer learning for nlp.
\newblock In \emph{International conference on machine learning}, pages 2790--2799. PMLR.

\bibitem[{Jiang et~al.(2023)Jiang, Sablayrolles, Mensch, Bamford, Chaplot, Casas, Bressand, Lengyel, Lample, Saulnier et~al.}]{jiang2023mistral}
Albert~Q Jiang, Alexandre Sablayrolles, Arthur Mensch, Chris Bamford, Devendra~Singh Chaplot, Diego de~las Casas, Florian Bressand, Gianna Lengyel, Guillaume Lample, Lucile Saulnier, et~al. 2023.
\newblock Mistral 7b.
\newblock \emph{arXiv preprint arXiv:2310.06825}.

\bibitem[{Jiang et~al.(2024)Jiang, Sablayrolles, Roux, Mensch, Savary, Bamford, Chaplot, Casas, Hanna, Bressand et~al.}]{jiang2024mixtral}
Albert~Q Jiang, Alexandre Sablayrolles, Antoine Roux, Arthur Mensch, Blanche Savary, Chris Bamford, Devendra~Singh Chaplot, Diego de~las Casas, Emma~Bou Hanna, Florian Bressand, et~al. 2024.
\newblock Mixtral of experts.
\newblock \emph{arXiv preprint arXiv:2401.04088}.

\bibitem[{Kim et~al.(2025)Kim, Yoo, Son, Patel, Agarwal, and Oh}]{kim2025benchhubunifiedbenchmarksuite}
Eunsu Kim, Haneul Yoo, Guijin Son, Hitesh Patel, Amit Agarwal, and Alice Oh. 2025.
\newblock \href {https://arxiv.org/abs/2506.00482} {Benchhub: A unified benchmark suite for holistic and customizable llm evaluation}.
\newblock \emph{Preprint}, arXiv:2506.00482.

\bibitem[{Kirk et~al.(2023)Kirk, Mediratta, Nalmpantis, Luketina, Hambro, Grefenstette, and Raileanu}]{kirk2023understanding}
Robert Kirk, Ishita Mediratta, Christoforos Nalmpantis, Jelena Luketina, Eric Hambro, Edward Grefenstette, and Roberta Raileanu. 2023.
\newblock Understanding the effects of rlhf on llm generalisation and diversity.
\newblock \emph{arXiv preprint arXiv:2310.06452}.

\bibitem[{Lauscher et~al.(2020)Lauscher, Ravishankar, Vuli{\'c}, and Glava{\v{s}}}]{lauscher2020zero}
Anne Lauscher, Vinit Ravishankar, Ivan Vuli{\'c}, and Goran Glava{\v{s}}. 2020.
\newblock From zero to hero: On the limitations of zero-shot cross-lingual transfer with multilingual transformers.
\newblock \emph{arXiv preprint arXiv:2005.00633}.

\bibitem[{Lewis et~al.(2019)Lewis, O{\u{g}}uz, Rinott, Riedel, and Schwenk}]{lewis2019mlqa}
Patrick Lewis, Barlas O{\u{g}}uz, Ruty Rinott, Sebastian Riedel, and Holger Schwenk. 2019.
\newblock Mlqa: Evaluating cross-lingual extractive question answering.
\newblock \emph{arXiv preprint arXiv:1910.07475}.

\bibitem[{Lewis et~al.(2020)Lewis, Perez, Piktus, Petroni, Karpukhin, Goyal, K{\"u}ttler, Lewis, Yih, Rockt{\"a}schel et~al.}]{lewis2020retrieval}
Patrick Lewis, Ethan Perez, Aleksandra Piktus, Fabio Petroni, Vladimir Karpukhin, Naman Goyal, Heinrich K{\"u}ttler, Mike Lewis, Wen-tau Yih, Tim Rockt{\"a}schel, et~al. 2020.
\newblock Retrieval-augmented generation for knowledge-intensive nlp tasks.
\newblock \emph{Advances in Neural Information Processing Systems}, 33:9459--9474.

\bibitem[{Li et~al.(2024{\natexlab{a}})Li, Alkhouli, Bonadiman, Pappas, and Mansour}]{li2024eliciting}
Bryan Li, Tamer Alkhouli, Daniele Bonadiman, Nikolaos Pappas, and Saab Mansour. 2024{\natexlab{a}}.
\newblock Eliciting better multilingual structured reasoning from llms through code.
\newblock \emph{arXiv preprint arXiv:2403.02567}.

\bibitem[{Li et~al.(2024{\natexlab{b}})Li, Haider, Luo, Agashe, and Callison-Burch}]{li2024bordirlines}
Bryan Li, Samar Haider, Fiona Luo, Adwait Agashe, and Chris Callison-Burch. 2024{\natexlab{b}}.
\newblock Bordirlines: A dataset for evaluating cross-lingual retrieval-augmented generation.
\newblock \emph{arXiv preprint arXiv:2410.01171}.

\bibitem[{Lin et~al.(2024)Lin, Wang, Liu, and Chen}]{lin2024crossin}
Geyu Lin, Bin Wang, Zhengyuan Liu, and Nancy~F Chen. 2024.
\newblock Crossin: An efficient instruction tuning approach for cross-lingual knowledge alignment.
\newblock \emph{arXiv preprint arXiv:2404.11932}.

\bibitem[{Lin et~al.(2021)Lin, Mihaylov, Artetxe, Wang, Chen, Simig, Ott, Goyal, Bhosale, Du et~al.}]{lin2021few}
Xi~Victoria Lin, Todor Mihaylov, Mikel Artetxe, Tianlu Wang, Shuohui Chen, Daniel Simig, Myle Ott, Naman Goyal, Shruti Bhosale, Jingfei Du, et~al. 2021.
\newblock Few-shot learning with multilingual language models.
\newblock \emph{arXiv preprint arXiv:2112.10668}.

\bibitem[{Lu et~al.(2024)Lu, Zhu, Li et~al.}]{lu2024llamax}
Yinquan Lu, Wenhao Zhu, Lei Li, et~al. 2024.
\newblock Llamax: Scaling linguistic horizons of llm by enhancing translation capabilities beyond 100 languages.
\newblock \emph{arXiv preprint arXiv:2407.05975}.

\bibitem[{Meghwani et~al.(2025)Meghwani, Agarwal, Pattnayak, Patel, and Panda}]{meghwani-etal-2025-hard}
Hansa Meghwani, Amit Agarwal, Priyaranjan Pattnayak, Hitesh~Laxmichand Patel, and Srikant Panda. 2025.
\newblock \href {https://doi.org/10.18653/v1/2025.acl-industry.72} {Hard negative mining for domain-specific retrieval in enterprise systems}.
\newblock In \emph{Proceedings of the 63rd Annual Meeting of the Association for Computational Linguistics (Volume 6: Industry Track)}, pages 1013--1026, Vienna, Austria. Association for Computational Linguistics.

\bibitem[{Nezhad and Agrawal(2024)}]{nezhad2024drives}
Sina~Bagheri Nezhad and Ameeta Agrawal. 2024.
\newblock What drives performance in multilingual language models?
\newblock \emph{arXiv preprint arXiv:2404.19159}.

\bibitem[{Nie et~al.(2022)Nie, Liang, Schmid, and Sch{\"u}tze}]{nie2022cross}
Ercong Nie, Sheng Liang, Helmut Schmid, and Hinrich Sch{\"u}tze. 2022.
\newblock Cross-lingual retrieval augmented prompt for low-resource languages.
\newblock \emph{arXiv preprint arXiv:2212.09651}.

\bibitem[{OpenAI(2024)}]{openai2024mmmlu}
OpenAI. 2024.
\newblock Multilingual massive multitask language understanding (mmmlu).
\newblock \url{https://huggingface.co/datasets/openai/MMMLU}.
\newblock Accessed: 2025-06-07.

\bibitem[{Panda et~al.(2025{\natexlab{a}})Panda, Agarwal, Nambirajan, and Pachauri}]{panda2025out}
Srikant Panda, Amit Agarwal, Gouttham Nambirajan, and Kulbhushan Pachauri. 2025{\natexlab{a}}.
\newblock Out of distribution element detection for information extraction.
\newblock US Patent App. 18/347,983.

\bibitem[{Panda et~al.(2025{\natexlab{b}})Panda, Agarwal, and Pachauri}]{panda2025dynamic}
Srikant Panda, Amit Agarwal, and Kulbhushan Pachauri. 2025{\natexlab{b}}.
\newblock Dynamic vocabularies for conditioning a language model for transforming natural language to a logical form.
\newblock US Patent App. 18/419,896.

\bibitem[{Panda et~al.(2025{\natexlab{c}})Panda, Agarwal, and Pachauri}]{panda2025techniques}
Srikant Panda, Amit Agarwal, and Kulbhushan Pachauri. 2025{\natexlab{c}}.
\newblock Techniques of information extraction for selection marks.
\newblock US Patent App. 18/240,344.

\bibitem[{Panda et~al.(2025{\natexlab{d}})Panda, Agarwal, and Patel}]{panda2025accessevalbenchmarkingdisabilitybias}
Srikant Panda, Amit Agarwal, and Hitesh~Laxmichand Patel. 2025{\natexlab{d}}.
\newblock \href {https://arxiv.org/abs/2509.22703} {Accesseval: Benchmarking disability bias in large language models}.
\newblock \emph{Preprint}, arXiv:2509.22703.

\bibitem[{Panda et~al.(2025{\natexlab{e}})Panda, Hari, Panda, Agarwal, and Patel}]{panda2025whosaskinginvestigatingbias}
Srikant Panda, Vishnu Hari, Kalpana Panda, Amit Agarwal, and Hitesh~Laxmichand Patel. 2025{\natexlab{e}}.
\newblock \href {https://arxiv.org/abs/2508.15831} {Who's asking? investigating bias through the lens of disability framed queries in llms}.
\newblock \emph{Preprint}, arXiv:2508.15831.

\bibitem[{Panda et~al.(2025{\natexlab{f}})Panda, Patel, Al-Khalifa, Agarwal, Al-Khalifa, and Al-Ghamdi}]{panda2025daiqauditingdemographicattribute}
Srikant Panda, Hitesh~Laxmichand Patel, Shahad Al-Khalifa, Amit Agarwal, Hend Al-Khalifa, and Sharefah Al-Ghamdi. 2025{\natexlab{f}}.
\newblock \href {https://arxiv.org/abs/2508.15830} {Daiq: Auditing demographic attribute inference from question in llms}.
\newblock \emph{Preprint}, arXiv:2508.15830.

\bibitem[{Paperno et~al.(2016)Paperno, Kruszewski, Lazaridou, Pham, Bernardi, Pezzelle, Baroni, Boleda, and Fern{\'a}ndez}]{paperno2016lambada}
Denis Paperno, Germ{\'a}n Kruszewski, Angeliki Lazaridou, Quan~Ngoc Pham, Raffaella Bernardi, Sandro Pezzelle, Marco Baroni, Gemma Boleda, and Raquel Fern{\'a}ndez. 2016.
\newblock The lambada dataset: Word prediction requiring a broad discourse context.
\newblock \emph{arXiv preprint arXiv:1606.06031}.

\bibitem[{Patel et~al.(2025{\natexlab{a}})Patel, Agarwal, Das, Kumar, Panda, Pattnayak, Rafi, Kumar, and Chae}]{patel2025sweeval}
Hitesh~Laxmichand Patel, Amit Agarwal, Arion Das, Bhargava Kumar, Srikant Panda, Priyaranjan Pattnayak, Taki~Hasan Rafi, Tejaswini Kumar, and Dong-Kyu Chae. 2025{\natexlab{a}}.
\newblock Sweeval: Do llms really swear? a safety benchmark for testing limits for enterprise use.
\newblock In \emph{Proceedings of the 2025 Conference of the Nations of the Americas Chapter of the Association for Computational Linguistics: Human Language Technologies (Volume 3: Industry Track)}, pages 558--582.

\bibitem[{Patel et~al.(2024{\natexlab{a}})Patel, Agarwal, Kumar, Gupta, and Pattnayak}]{patel2024llmbarcodesgeneratingdiverse}
Hitesh~Laxmichand Patel, Amit Agarwal, Bhargava Kumar, Karan Gupta, and Priyaranjan Pattnayak. 2024{\natexlab{a}}.
\newblock \href {https://arxiv.org/abs/2411.14962} {Llm for barcodes: Generating diverse synthetic data for identity documents}.
\newblock \emph{Preprint}, arXiv:2411.14962.

\bibitem[{Patel et~al.(2024{\natexlab{b}})Patel, Agarwal, Kumar, Gupta, and Pattnayak}]{patel2024llm}
Hitesh~Laxmichand Patel, Amit Agarwal, Bhargava Kumar, Karan Gupta, and Priyaranjan Pattnayak. 2024{\natexlab{b}}.
\newblock Llm for barcodes: Generating diverse synthetic data for identity documents.
\newblock \emph{arXiv preprint arXiv:2411.14962}.

\bibitem[{Patel et~al.(2025{\natexlab{b}})Patel, Agarwal, Panda, Meghwani, Dua, Li, Sheng, Ravi, and Roth}]{patel2025pcrimeasuringcontextrobustness}
Hitesh~Laxmichand Patel, Amit Agarwal, Srikant Panda, Hansa Meghwani, Karan Dua, Paul Li, Tao Sheng, Sujith Ravi, and Dan Roth. 2025{\natexlab{b}}.
\newblock \href {https://arxiv.org/abs/2509.23879} {Pcri: Measuring context robustness in multimodal models for enterprise applications}.
\newblock \emph{Preprint}, arXiv:2509.23879.

\bibitem[{Pattnayak et~al.(2025{\natexlab{a}})Pattnayak, Agarwal, Meghwani, Patel, and Panda}]{pattnayak2025hybrid}
Priyaranjan Pattnayak, Amit Agarwal, Hansa Meghwani, Hitesh~Laxmichand Patel, and Srikant Panda. 2025{\natexlab{a}}.
\newblock Hybrid ai for responsive multi-turn online conversations with novel dynamic routing and feedback adaptation.
\newblock In \emph{Proceedings of the 4th International Workshop on Knowledge-Augmented Methods for Natural Language Processing}, pages 215--229.

\bibitem[{Pattnayak et~al.(2025{\natexlab{b}})Pattnayak, Patel, and Agarwal}]{pattnayak2025tokenizationmattersimprovingzeroshot}
Priyaranjan Pattnayak, Hitesh~Laxmichand Patel, and Amit Agarwal. 2025{\natexlab{b}}.
\newblock \href {https://arxiv.org/abs/2504.16977} {Tokenization matters: Improving zero-shot ner for indic languages}.
\newblock \emph{Preprint}, arXiv:2504.16977.

\bibitem[{Pattnayak et~al.(2025{\natexlab{c}})Pattnayak, Patel, Agarwal, Kumar, Panda, and Kumar}]{pattnayak2025clinicalqa20multitask}
Priyaranjan Pattnayak, Hitesh~Laxmichand Patel, Amit Agarwal, Bhargava Kumar, Srikant Panda, and Tejaswini Kumar. 2025{\natexlab{c}}.
\newblock \href {https://arxiv.org/abs/2502.13108} {Clinical qa 2.0: Multi-task learning for answer extraction and categorization}.
\newblock \emph{Preprint}, arXiv:2502.13108.

\bibitem[{Pattnayak et~al.(2024)Pattnayak, Patel, Kumar, Agarwal, Banerjee, Panda, and Kumar}]{pattnayak2024survey}
Priyaranjan Pattnayak, Hitesh~Laxmichand Patel, Bhargava Kumar, Amit Agarwal, Ishan Banerjee, Srikant Panda, and Tejaswini Kumar. 2024.
\newblock Survey of large multimodal model datasets, application categories and taxonomy.
\newblock \emph{arXiv preprint arXiv:2412.17759}.

\bibitem[{Pfeiffer et~al.(2020)Pfeiffer, Kamath, R{\"u}ckl{\'e}, Cho, and Gurevych}]{pfeiffer2020adapterfusion}
Jonas Pfeiffer, Aishwarya Kamath, Andreas R{\"u}ckl{\'e}, Kyunghyun Cho, and Iryna Gurevych. 2020.
\newblock Adapterfusion: Non-destructive task composition for transfer learning.
\newblock \emph{arXiv preprint arXiv:2005.00247}.

\bibitem[{Pires(2019)}]{pires2019multilingual}
T~Pires. 2019.
\newblock How multilingual is multilingual bert.
\newblock \emph{arXiv preprint arXiv:1906.01502}.

\bibitem[{Rafailov et~al.(2024)Rafailov, Sharma, Mitchell, Manning, Ermon, and Finn}]{rafailov2024direct}
Rafael Rafailov, Archit Sharma, Eric Mitchell, Christopher~D Manning, Stefano Ermon, and Chelsea Finn. 2024.
\newblock Direct preference optimization: Your language model is secretly a reward model.
\newblock \emph{Advances in Neural Information Processing Systems}, 36.

\bibitem[{Shaham et~al.(2024)Shaham, Herzig, Aharoni, Szpektor, Tsarfaty, and Eyal}]{shaham2024multilingual}
Uri Shaham, Jonathan Herzig, Roee Aharoni, Idan Szpektor, Reut Tsarfaty, and Matan Eyal. 2024.
\newblock Multilingual instruction tuning with just a pinch of multilinguality.
\newblock \emph{arXiv preprint arXiv:2401.01854}.

\bibitem[{She et~al.(2024)She, Zou, Huang, Zhu, Liu, Geng, and Chen}]{she2024mapo}
Shuaijie She, Wei Zou, Shujian Huang, Wenhao Zhu, Xiang Liu, Xiang Geng, and Jiajun Chen. 2024.
\newblock Mapo: Advancing multilingual reasoning through multilingual alignment-as-preference optimization.
\newblock \emph{arXiv preprint arXiv:2401.06838}.

\bibitem[{Shi et~al.(2022)Shi, Suzgun, Freitag, Wang, Srivats, Vosoughi, Chung, Tay, Ruder, Zhou, Das, and Wei}]{shi2022language}
Freda Shi, Mirac Suzgun, Markus Freitag, Xuezhi Wang, Suraj Srivats, Soroush Vosoughi, Hyung~Won Chung, Yi~Tay, Sebastian Ruder, Denny Zhou, Dipanjan Das, and Jason Wei. 2022.
\newblock \href {https://arxiv.org/abs/2210.03057} {Language models are multilingual chain-of-thought reasoners}.
\newblock \emph{Preprint}, arXiv:2210.03057.

\bibitem[{Singh(2021)}]{Singh2021}
Jyotika Singh. 2021.
\newblock \href {https://doi.org/10.25080/majora-1b6fd038-009} {Social media analysis using natural language processing techniques}.
\newblock In \emph{Proceedings of the 20th Python in Science Conference}, SciPy, page 74–80. SciPy.

\bibitem[{Singh(2022)}]{Singh2022}
Jyotika Singh. 2022.
\newblock \href {https://doi.org/10.25080/majora-212e5952-017} {pyaudioprocessing: Audio processing, feature extraction, and machine learning modeling}.
\newblock In \emph{Proceedings of the 21st Python in Science Conference}, SciPy, page 152–158. SciPy.

\bibitem[{Singh(2023)}]{Singh2023}
Jyotika Singh. 2023.
\newblock \href {https://doi.org/10.1201/9781003264774} {\emph{Natural Language Processing in the Real World: Text Processing, Analytics, and Classification}}.
\newblock Chapman and Hall/CRC.

\bibitem[{Son et~al.(2025)Son, Yang, Patel, Agarwal, Ko, Lim, Panda, Kim, Drolia, Choi, Lee, and Yu}]{son2025pushingmultilingualreasoningmodels}
Guijin Son, Donghun Yang, Hitesh~Laxmichand Patel, Amit Agarwal, Hyunwoo Ko, Chanuk Lim, Srikant Panda, Minhyuk Kim, Nikunj Drolia, Dasol Choi, Kyong-Ha Lee, and Youngjae Yu. 2025.
\newblock \href {https://arxiv.org/abs/2510.04230} {Pushing on multilingual reasoning models with language-mixed chain-of-thought}.
\newblock \emph{Preprint}, arXiv:2510.04230.

\bibitem[{Tarunesh et~al.(2021)Tarunesh, Khyalia, Kumar, Ramakrishnan, and Jyothi}]{tarunesh2021meta}
Ishan Tarunesh, Sushil Khyalia, Vishwajeet Kumar, Ganesh Ramakrishnan, and Preethi Jyothi. 2021.
\newblock Meta-learning for effective multi-task and multilingual modelling.
\newblock \emph{arXiv preprint arXiv:2101.10368}.

\bibitem[{Thakur et~al.(2023)Thakur, Bonifacio, Zhang, Ogundepo, Kamalloo, Alfonso-Hermelo, Li, Liu, Chen, Rezagholizadeh et~al.}]{thakur2023nomiracl}
Nandan Thakur, Luiz Bonifacio, Xinyu Zhang, Odunayo Ogundepo, Ehsan Kamalloo, David Alfonso-Hermelo, Xiaoguang Li, Qun Liu, Boxing Chen, Mehdi Rezagholizadeh, et~al. 2023.
\newblock Nomiracl: Knowing when you don't know for robust multilingual retrieval-augmented generation.
\newblock \emph{arXiv preprint arXiv:2312.11361}.

\bibitem[{Thomas et~al.(2025)Thomas, Agarwal, Jana, and Pachauri}]{thomas2025model}
Edwin Thomas, Amit Agarwal, Sandeep Jana, and Kulbhushan Pachauri. 2025.
\newblock Model augmentation framework for domain assisted continual learning in deep learning.
\newblock US Patent App. 18/406,905.

\bibitem[{Touvron et~al.(2023)Touvron, Lavril, Izacard, Martinet, Lachaux, Lacroix, Rozi{\`e}re, Goyal, Hambro, Azhar et~al.}]{touvron2023llama}
Hugo Touvron, Thibaut Lavril, Gautier Izacard, Xavier Martinet, Marie-Anne Lachaux, Timoth{\'e}e Lacroix, Baptiste Rozi{\`e}re, Naman Goyal, Eric Hambro, Faisal Azhar, et~al. 2023.
\newblock Llama: Open and efficient foundation language models.
\newblock \emph{arXiv preprint arXiv:2302.13971}.

\bibitem[{Wasi et~al.(2025)Wasi, Islam, Rahman, Yeasmin, Agarwal, Patel, Rafi, and Chae}]{wasi2025legalmind}
Azmine~Toushik Wasi, Mst~Rafia Islam, Abdur Rahman, Tawfia Yeasmin, Amit Agarwal, Hitesh~Laxmichand Patel, Taki~Hasan Rafi, and Dong-Kyu Chae. 2025.
\newblock Legalmind: An intelligent solution for legal document analysis with user-centric ui and ai-driven capabilities in mobile devices.
\newblock In \emph{Companion Publication of the 2025 Conference on Computer-Supported Cooperative Work and Social Computing}, pages 410--414.

\bibitem[{Xu et~al.(2024)Xu, Hu, Zhao, Qiu, Ye, and Gu}]{xu2024survey}
Yuemei Xu, Ling Hu, Jiayi Zhao, Zihan Qiu, Yuqi Ye, and Hanwen Gu. 2024.
\newblock A survey on multilingual large language models: Corpora, alignment, and bias.
\newblock \emph{arXiv preprint arXiv:2404.00929}.

\bibitem[{Xue et~al.(2021)Xue, Constant, Roberts, Kale, Al-Rfou, Siddhant, Barua, and Raffel}]{xue-etal-2021-mt5}
Linting Xue, Noah Constant, Adam Roberts, Mihir Kale, Rami Al-Rfou, Aditya Siddhant, Aditya Barua, and Colin Raffel. 2021.
\newblock \href {https://doi.org/10.18653/v1/2021.naacl-main.41} {m{T}5: A massively multilingual pre-trained text-to-text transformer}.
\newblock In \emph{Proceedings of the 2021 Conference of the North American Chapter of the Association for Computational Linguistics: Human Language Technologies}, pages 483--498, Online. Association for Computational Linguistics.

\bibitem[{Yang et~al.(2024)Yang, Wu, Wang, Zong, and Zhang}]{yang2024language}
Wen Yang, Junhong Wu, Chen Wang, Chengqing Zong, and Jiajun Zhang. 2024.
\newblock Language imbalance driven rewarding for multilingual self-improving.
\newblock \emph{arXiv preprint arXiv:2410.08964}.

\bibitem[{Yang et~al.(2019)Yang, Zhang, Tar, and Baldridge}]{yang-etal-2019-paws}
Yinfei Yang, Yuan Zhang, Chris Tar, and Jason Baldridge. 2019.
\newblock \href {https://doi.org/10.18653/v1/D19-1382} {{PAWS}-{X}: A cross-lingual adversarial dataset for paraphrase identification}.
\newblock In \emph{Proceedings of the 2019 Conference on Empirical Methods in Natural Language Processing and the 9th International Joint Conference on Natural Language Processing (EMNLP-IJCNLP)}, pages 3687--3692, Hong Kong, China. Association for Computational Linguistics.

\bibitem[{Yin et~al.(2024)Yin, Wan, Shen, Patel, Li, Gu, and Xiong}]{yin2024continuous}
Nan Yin, Mengzhu Wan, Li~Shen, Hitesh~Laxmichand Patel, Baopu Li, Bin Gu, and Huan Xiong. 2024.
\newblock Continuous spiking graph neural networks.
\newblock \emph{arXiv preprint arXiv:2404.01897}.

\bibitem[{Zeng et~al.(2025)Zeng, Han, Chen, and Yu}]{zeng-etal-2025-converging}
Hongchuan Zeng, Senyu Han, Lu~Chen, and Kai Yu. 2025.
\newblock \href {https://aclanthology.org/2025.coling-main.707/} {Converging to a lingua franca: Evolution of linguistic regions and semantics alignment in multilingual large language models}.
\newblock In \emph{Proceedings of the 31st International Conference on Computational Linguistics}, pages 10602--10617, Abu Dhabi, UAE. Association for Computational Linguistics.

\bibitem[{Zhang et~al.(2024)Zhang, Jin, Huang, Zhang, and Wei}]{zhang2024respond}
Liang Zhang, Qin Jin, Haoyang Huang, Dongdong Zhang, and Furu Wei. 2024.
\newblock Respond in my language: Mitigating language inconsistency in response generation based on large language models.
\newblock In \emph{Proceedings of the 62nd Annual Meeting of the Association for Computational Linguistics (Volume 1: Long Papers)}, pages 4177--4192.

\bibitem[{Zhang et~al.(2023)Zhang, Cahyawijaya, Cruz, Winata, and Aji}]{zhang-etal-2023-multilingual}
Ruochen Zhang, Samuel Cahyawijaya, Jan Christian~Blaise Cruz, Genta Winata, and Alham~Fikri Aji. 2023.
\newblock \href {https://doi.org/10.18653/v1/2023.emnlp-main.774} {Multilingual large language models are not (yet) code-switchers}.
\newblock In \emph{Proceedings of the 2023 Conference on Empirical Methods in Natural Language Processing}, pages 12567--12582, Singapore. Association for Computational Linguistics.

\bibitem[{Zhao et~al.(2024)Zhao, Zhang, Chen, Kawaguchi, and Bing}]{zhao2024large}
Yiran Zhao, Wenxuan Zhang, Guizhen Chen, Kenji Kawaguchi, and Lidong Bing. 2024.
\newblock How do large language models handle multilingualism?
\newblock \emph{arXiv preprint arXiv:2402.18815}.

\bibitem[{Zhu et~al.(2024)Zhu, Liu, Dong, Xu, Huang, Kong, Chen, and Li}]{zhu-etal-2024-multilingual}
Wenhao Zhu, Hongyi Liu, Qingxiu Dong, Jingjing Xu, Shujian Huang, Lingpeng Kong, Jiajun Chen, and Lei Li. 2024.
\newblock \href {https://doi.org/10.18653/v1/2024.findings-naacl.176} {Multilingual machine translation with large language models: Empirical results and analysis}.
\newblock In \emph{Findings of the Association for Computational Linguistics: NAACL 2024}, pages 2765--2781, Mexico City, Mexico. Association for Computational Linguistics.

\bibitem[{Zhu et~al.(2023)Zhu, Lv, Dong, Yuan, Xu, Huang, Kong, Chen, and Li}]{zhu2023extrapolating}
Wenhao Zhu, Yunzhe Lv, Qingxiu Dong, Fei Yuan, Jingjing Xu, Shujian Huang, Lingpeng Kong, Jiajun Chen, and Lei Li. 2023.
\newblock Extrapolating large language models to non-english by aligning languages.
\newblock \emph{arXiv preprint arXiv:2308.04948}.

\end{thebibliography}

\appendix

\clearpage

\section{Appendix}
\label{sec:appendix}

\subsection{Extended Related Works}
\label{sec:appendix_related}

\paragraph{Multilingual Consistency in Large Language Models.}  
Recent advances in multilingual LLMs aim to extend the capabilities of LLMs to a broad spectrum of languages, including low-resource and non-Latin scripts enabling systems like RAG. \citet{nezhad2024drives} analyzed factors influencing LLM performance across over 200 languages, showing that pretraining data size predominantly governs seen-language performance, while script type and language family are key for unseen languages. This underscores the importance of balanced and diverse pretraining data to achieve consistent multilingual performance. \citet{zhao2024large} studied the internal mechanisms of multilingual LLMs and revealed a prevalent “English-centric” reasoning phenomenon, where models often perform intermediate reasoning steps in English before generating outputs in the target language. This behavior contributes to cross-lingual inconsistencies across enterprise applications with global customer base, motivating methods targeting internal alignment.



\textbf{Cross-Lingual Transfer and Preference Alignment.}  
Cross-lingual knowledge transfer remains essential for robust multilingual LLMs \cite{son2025pushingmultilingualreasoningmodels}. \citet{zhu-etal-2024-multilingual} assessed LLM translation capabilities, noting that although models like GPT-4 improve high-resource language performance, they still lag behind specialized supervised systems in low-resource scenarios. Fine-tuning has shown promise over in-context learning, particularly for low-resource languages \citep{asai2024buffet}. Additionally, \citet{shaham2024multilingual} proposed minimal multilingual instruction tuning, demonstrating that even limited multilingual data can significantly enhance cross-lingual generalization and instruction following on unseen languages.  

Recent research has expanded the landscape of multilingual alignment and consistency in LLMs. MAPO~\citep{she2024mapo} employs multi-stage, translation-based preference optimization to align multilingual LLMs, while Language Imbalance Driven Rewarding~\citep{yang2024language} addresses training imbalances to improve fairness and accuracy. Zeng et al.~\citep{zeng-etal-2025-converging} show that LLMs naturally develop a shared internal representation during multilingual training, supporting the idea that intrinsic multilingual reasoning can be optimized through targeted alignment.

Several very recent works propose complementary strategies. ~\citealp{zhu2023extrapolating} and ~\citealp{lin2024crossin} explore cross-lingual instruction tuning, leveraging explicit translation tasks and mixed datasets to boost non-English and cross-lingual abilities, with a particular emphasis on data allocation, extraction and scaling \cite{panda2025dynamic,}. \citep{dang2024rlhf} and \citep{she2024mapo} advance preference-based optimization for multilingual LLMs, using reward models, synthetic or human preference data~\cite{patel2024llmbarcodesgeneratingdiverse,panda2025techniques,dua2025flexdocparameterizedsamplingdiverse}, and in some cases external translation models to align non-English outputs with English anchors.

In contrast, our work achieves multilingual alignment by directly leveraging semantically equivalent multilingual data within each training batch—\textbf{without requiring explicit translation supervision, reward models, or preference data}. This simple, batch-level alignment enables robust multilingual consistency and scalability for industry deployment. Our approach is orthogonal and complementary to recent advances, providing a more straightforward and resource-efficient alternative for practical multilingual LLM alignment.

\paragraph{Multilingual Consistency in RAG.}  
Ensuring consistent multilingual reasoning and generation within RAG systems is a challenging problem, critical to enterprise applications which serve the same information across different languages to a global user-base \cite{wasi2025legalmind,pattnayak2025hybrid,meghwani-etal-2025-hard,Singh2023,Singh2021}. The NoMIRACL benchmark \citep{thakur2023nomiracl} was introduced to rigorously evaluate RAG model's ability to maintain factual consistency across languages by pairing queries with relevant and irrelevant documents in multiple languages. ~\citep{zhang2024respond} focus on penalizing English responses to non-English prompts, addressing output language consistency without using multilingual data for fine-tuning, while effective, challenges remain due to differences in topic distributions and retrieval noise across languages and data types \cite{dua-etal-2025-speechweave,Singh2022}. \citet{zhang-etal-2023-multilingual} further demonstrated challenges faced by multilingual LLMs with code-switched inputs, emphasizing the need for specialized training to handle linguistically complex scenarios. 

Recent benchmarks like BUFFET \citep{asai2024buffet}, MMMLU \cite{openai2024mmmlu}, MGSM \cite{cobbe2021training,shi2022language}, BenchHub \cite{kim2025benchhubunifiedbenchmarksuite}, SEA-VL \cite{cahyawijaya-etal-2025-crowdsource}, LAMBADA Multilingual \cite{eleutherai2024lambadamultilingual,paperno2016lambada}, PAWS-X \cite{yang-etal-2019-paws} and xSTREET \citep{li2024eliciting} reveal that LLMs still struggle to maintain consistent reasoning across languages, indicating that fundamental challenges remain; motivating the need for alignment-focused methods. 

Our approach advances this by proposing a batch-wise alignment strategy using semantically equivalent multilingual data within each training batch, enabling direct alignment of reasoning and generation across languages during fine-tuning without dependence on external translation, rewards models or retrieval components.

\paragraph{Extensive Pre-training for Improved Translation.}
\citealp{lu2024llamax} addressed gaps in multilingual capabilities by developing LLaMAX, a language model designed for enhanced translation across over 100 languages. Their approach involved extensive multilingual pretraining with data augmentation and vocabulary expansion, leading to significant improvements in both high-resource and low-resource translation tasks. This study demonstrates the importance of robust multilingual pretraining to boost LLMs’ language generation capabilities.

\paragraph{Multilingual Pretraining and LLMs.}
Multilingual pretraining has been central to improving the performance of LLMs across different languages, providing a means to develop models that generalize effectively across diverse linguistic domains. Early models like mBERT \cite{devlin2018bert, pires2019multilingual} laid the groundwork by showing that shared embeddings could facilitate cross-lingual transfer. However, models such as XLM-R \cite{conneau2019unsupervised}, with its large-scale pretraining on 2 terabytes of CommonCrawl data across 100 languages, demonstrated significant improvements in multilingual understanding, outperforming mBERT on benchmarks such as XNLI \cite{conneau2018xnli} and MLQA \cite{lewis2019mlqa}.

mT5 \cite{xue-etal-2021-mt5} further improved on these models by unifying the text-to-text framework for multilingual tasks, supporting a wide range of NLP tasks across 101 languages. 
LaBSE v2, released in 2023, builds on LaBSE \cite{feng2020language} and employs enhanced language-agnostic training objectives to achieve superior cross-lingual sentence embeddings. This model is particularly well-suited for sentence-level tasks like retrieval and paraphrasing across 200 languages, including many low-resource languages, offering improvements over prior models by incorporating alignment techniques for better sentence representation. Furthermore, XGLM \cite{lin2021few}, an autoregressive language model trained on massive multilingual data, has become a recent state-of-the-art approach for multilingual tasks, using a probabilistic approach to model language sequences.


\noindent\textbf{Multilingual Fine-tuning Strategies.}
Fine-tuning multilingual models for specific tasks like document understanding \cite{agarwal-etal-2025-fs,yin2024continuous,patel2024llm,panda2025out}, video understanding \cite{agarwal2021evaluate,agarwal-etal-2025-mvtamperbench,thomas2025model}, clinical trials \cite{pattnayak2025clinicalqa20multitask}, and accessible applications  \cite{panda2025accessevalbenchmarkingdisabilitybias,panda2025whosaskinginvestigatingbias,panda2025daiqauditingdemographicattribute} has evolved rapidly in recent years along with applications requiring both multi-modal and multilingual capabilities from models \cite{pattnayak2024survey}. A key advancement is the use of adapter layers \cite{houlsby2019parameter}, which enable task-specific fine-tuning without modifying the entire model, resulting in reduced computational overhead. AdapterFusion \cite{pfeiffer2020adapterfusion} further extends this by fusing knowledge from multiple task-specific adapters, allowing for better transfer across tasks and languages. 

A growing body of research has shown that fine-tuning on typologically similar languages can improve performance on low-resource languages \cite{pattnayak2025tokenizationmattersimprovingzeroshot}. \citealp{lauscher2020zero} demonstrated the effectiveness of cross-lingual transfer learning through multilingual fine-tuning, showing that models fine-tuned on a high-resource language (e.g., Spanish) could improve performance on a related low-resource language (e.g., Catalan). 

Recent studies have also explored multilingual multi-task learning (MTL) as a fine-tuning strategy. \citealp{tarunesh2021meta} proposed a new multitask learning approach that fine-tunes multilingual models across several NLP tasks simultaneously, including machine translation, summarization, and question answering. Their results showed that training models on multiple tasks simultaneously can lead to performance improvements in individual tasks, particularly for low-resource languages.



\textbf{Alignment Strategies for Multilingual Models.}  
Beyond retrieval, aligning multilingual models internally has gained interest, which we further study in Appendix \ref{sec:appendix_related}. Recent methods explore multi-task and preference-based fine-tuning to reduce language biases and improve consistency \citep{patel2025sweeval,rafailov2024direct, hong2024reference,she2024mapo,yang2024language,zeng-etal-2025-converging}. Our approach advances this by proposing a batch-wise alignment strategy using semantically equivalent multilingual data within each training batch, enabling direct alignment of generation across languages during fine-tuning without dependence on external translation or retrieval components.

\subsection{Preliminaries}
\label{sec:appendix_prelim}
\noindent\textbf{Direct Preference Optimization (DPO).} DPO \cite{rafailov2024direct} is an implicit reward model for RLHF \cite{bai2022training,kirk2023understanding}  that ensures optimal policy to solve RLHF with a standard classification loss without needing RL to align human preference. Intuitively, it ensures a robust LLM fine-tuning without excessive hyperparameter tuning in a RL-free training environment, aligning LLMs with human preferences. Likewise, DPO is a straightforward method that can maximize the reward with KL-divergence constraint and find optimal policy based on a binary cross-entropy objective that enables simple training paradigm without additional RL overhead. 

\noindent\textbf{Odds-Ratio Preference Optimization (ORPO).} OPRO \cite{hong2024orpo} is a supervised fine-tuning (SFT) based preference alignment optimization method. Unlike DPO and RLHF, which typically need a separate reference model with SFT, whereas ORPO does not require a reference model by weakly penalizing undesired generations and pass strong adaptation signal for selective responses with an odd-ratio term.


\subsection{Extended Experimental Setup}
\label{sec:appendix_setup}
\noindent \textbf{Evaluation \& Dataset.} We first evaluate pretrained LLM performance \& instruction tuned LLM using standard DPO/ORPO training without batch modification as a control. We then apply our modified batch construction, where each batch contains different language versions of the same topic. The dataset is designed to provide consistent context across languages, and is split 70\% for training, 
and 30\% for testing, ensuring balanced representation across topics and business domains.

\noindent\textbf{Evaluation Models.} We evaluate eight open-sourced models,including three versions of the Mistral-7B \cite{jiang2023mistral} model, specifically {v0.1, v0.2, v0.3}, Mixtral-8x22B \cite{jiang2024mixtral} model, and four models from the Llama family \cite{touvron2023llama,dubey2024llama}, specifically one 8B and one 70B model from versions 3 and 3.1, respectively. Unless otherwise stated, all LLMs referenced in this work are their instruct/tuned versions.

\noindent\textbf{Training Pipeline.}\label{sec:appendix_hardware} Figure \ref{fig:training_pipeline} highlights the training paradigm for the proposed batch-alignment training. The LLMs are fine-tuned using both standard and modified DPO/ORPO training methods. We use a learning rate of 5e-6, batch size of 7, and a maximum sequence length of 4096 tokens. Training is performed for 3 epochs using mixed-precision on 4 NVIDIA A100 GPUs to reduce memory footprint and improve computational efficiency.

\subsubsection{Metrics for Evaluation}

We use Accuracy as the primary evaluation metric, as it effectively captures both true positives and true negatives, reflecting the model’s ability to reason consistently across multiple languages. Since our setup ensures that the model is expected to generate predefined responses (i.e., "Yes, the answer is present" or "I don't know"), we use exact match accuracy to determine correctness against the ground truth.
Unlike the NoMIRACL benchmark, which evaluates Hallucination Rate and Error Rate, we argue that Accuracy is a more comprehensive measure of reasoning consistency in multilingual settings. Hallucination Rate focuses solely on incorrect additions of information, while Error Rate does not differentiate between systematic generation errors and retrieval-induced inconsistencies. In contrast, Accuracy reflects the true end-to-end reasoning capability of the model across different languages, making it the most direct measure of multilingual consistency.


 While accuracy is well-suited to our binary QA formulation, we acknowledge the value of complementary metrics (such as semantic similarity or human evaluation) for more nuanced assessments of multilingual alignment across diverse textual and visual tasks \cite{agarwal2025rciscoreevaluatingglobal,patel2025pcrimeasuringcontextrobustness}.


\subsubsection{Prompt Design}
\label{sec:appendix_prompt}
Following NoMiracl, we use a structured prompt to query the LLM during training and evaluation. The prompt is designed to provide a consistent query format across all languages:


\begin{table*}[!htb]
    \centering
    \scalebox{0.77}{\begin{tabular}{lccccccccc}
        \toprule
        \multirow{2}{*}{\textbf{Models}} & \multicolumn{8}{c}{\textbf{Baseline}} \\
        \cmidrule(lr){2-9}
         & \textbf{English} & \textbf{Arabic} & \textbf{Spanish} & \textbf{French} & \textbf{Japanese} & \textbf{Portuguese} & \textbf{Chinese} & \textbf{Rest (avg.)}\\
        \midrule
      Mistral-7B-Instruct-v0.1    & 56.0        & 7.0         & 51.4        & 50.6        & 37.6        & 51.4        & 49.8        & 41.3               \\
\hline
Mistral-7B-Instruct-v0.2    & 64.0        & 57.2        & 63.0        & 61.0        & 61.2        & 63.6        & 61.6        & 61.3               \\
\hline
Mistral-7B-Instruct-v0.3    & 70.4        & 55.6        & 65.8        & 66.8        & 62.8        & 70.6        & 62.6        & 64.0               \\
\hline
Mixtral-8x22B-Instruct-v0.1 & 74.2        & 57.2        & 71.2        & 71.4        & 65.4        & 69.8        & 62.8        & 66.3               \\
\hline
Llama-3-8B-Instruct    & 56.0        & 50.6        & 57.4        & 44.0        & 49.6        & 54.8        & 54.4        & 51.8               \\
\hline
Llama-3-70B-Instruct   & 65.8        & 60.8        & 64.4        & 60.9        & 63.2        & 63.4        & 63.8        & 62.8               \\
\hline
Llama-3.1-8B-Instruct  & 59.2        & 24.8        & 38.4        & 6.8         & 49.4        & 29.0        & 50.4        & 33.1               \\
\hline
Llama-3.1-70B-Instruct & 73.6      & 45.6        & 19.6        & 50.4        & 65.8        & 22.6        & 65.4        & 44.9  \\ 
        \midrule
    \end{tabular}}
    \caption{Accuracy results of pre-trained LLMs for English and six non-English languages. The 'Rest (avg.)' column shows the average accuracy for non-English languages, highlighting a performance gap with English.}\label{tab1}
\end{table*}

\begin{verbatim}
I will give you a question and
several contexts containing 
information about the question. 
Read the contexts carefully. 
If any of the contexts answers 
the question, respond as either 
“Yes, answer is present” 
or “I don’t know”:
QUESTION: {query}
CONTEXTS:
[1] {Passage title}: {Passage text} 
[2] {Passage title}: {Passage text}
...
[10] {Passage title}: {Passage text}
OUTPUT:
\end{verbatim}

The prompt in our dataset is designed to simulate a typical RAG system, where each passage is retrieved from a knowledge source. During training and evaluation, every passage addresses the same topic but is presented in a different language.

\subsubsection{Design Rationale}
\label{sec:appendix_design}
We use a single relevant document per query, in line with the NoMIRACL framework, to ensure a controlled and interpretable evaluation of multilingual consistency, free from confounding effects of retrieval quality or relevance ranking. The binary classification setup enables straightforward exact-match evaluation, providing a robust measure of cross-lingual reasoning alignment.

We also standardize on English prompts for experimental clarity and reproducibility. However, our approach can be readily adapted to use other high-resource languages as alignment anchors, making it extensible to different multilingual scenarios.

\begin{table*}[!htb]
    \centering
    \scalebox{0.77}{\begin{tabular}{lccccccccc}
        \toprule
        \multirow{2}{*}{\textbf{Models}} & \multicolumn{8}{c}{\textbf{DPO Finetuning (Default)}} \\
        \cmidrule(lr){2-9}
         & \textbf{English} & \textbf{Arabic} & \textbf{Spanish} & \textbf{French} & \textbf{Japanese} & \textbf{Portuguese} & \textbf{Chinese} & \textbf{Rest (avg.)}\\
        \midrule
       Mistral-7B-Instruct-v0.1    & 60.3        & 27.2        & 54.6        & 54.6        & 44.4        & 57.7        & 52.5        & 48.5               \\
\hline
Mistral-7B-Instruct-v0.2    & 73.5        & 61.2        & 66.4        & 65.8        & 64.2        & 66.2        & 67.6        & 65.2               \\
\hline
Mistral-7B-Instruct-v0.3    & 76.3        & 62.3        & 66.8        & 67.4        & 65.3        & 71.5        & 68.9        & 67.0               \\
\hline
Mixtral-8x22B-Instruct-v0.1 & 79.3        & 63.4        & 72.9        & 72.6        & 67.9        & 72.1        & 69.6        & 69.8               \\
\hline
Llama-3-8B-Instruct    & 65.8        & 61.9        & 58.3        & 55.4        & 60.8        & 55.9        & 55.8        & 58.0               \\
\hline
Llama-3-70B-Instruct   & 73.6        & 64.8        & 64.4        & 69.0        & 67.2        & 61.8        & 68.8        & 66.0               \\
\hline
Llama-3.1-8B-Instruct  & 68.2        & 32.3        & 43.4        & 21.5        & 52.3        & 34.3        & 52.3        & 39.4               \\
\hline
Llama-3.1-70B-Instruct & 78.8        & 48.6        & 29.4        & 54.5        & 68.9        & 32.3        & 69.5        & 50.5   \\
        \midrule
        \multirow{2}{*}{\textbf{Models}} & \multicolumn{8}{c}{\textbf{DPO + Ours (Batch Size = 7)}} \\
        \cmidrule(lr){2-9}
        & \textbf{English} & \textbf{Arabic} & \textbf{Spanish} & \textbf{French} & \textbf{Japanese} & \textbf{Portuguese} & \textbf{Chinese} & \textbf{Rest (avg.)}\\
        \midrule
       Mistral-7B-Instruct-v0.1    & 60.3        & 45.6\scriptsize{(+18.4)}   & 58.6\scriptsize{(+4.0)}       & 57.6\scriptsize{(+3.0)}        & 55.6\scriptsize{(+11.2)}        & 57.7\scriptsize{(+0.0)}        & 54.9\scriptsize{(+2.4)}      & 55.0\scriptsize{(+6.5)}              \\
\hline
Mistral-7B-Instruct-v0.2    & 73.5        & 68.6\scriptsize{(+7.4)}        & 70.2\scriptsize{(+3.8)}        & 69.6\scriptsize{(+3.8)}        & 69.5\scriptsize{(+5.3)}        & 72.2\scriptsize{(+6.0)}        & 71.6\scriptsize{(+4.0)}        & 70.3\scriptsize{(+5.1)}               \\
\hline
Mistral-7B-Instruct-v0.3    & 76.3        & 71.6\scriptsize{(+9.3)}        & 72.6\scriptsize{(+5.8)}        & 73.5\scriptsize{(+6.1)}        & 71.2\scriptsize{(+5.9)}        & 74.5\scriptsize{(+3.0)}        & 72.9\scriptsize{(+4.0)}        & 72.7\scriptsize{(+5.7)}               \\
\hline
Mixtral-8x22B-Instruct-v0.1 & 79.3        & 72.5\scriptsize{(+9.1)}        & 75.6\scriptsize{(+2.7)}        & 74.6\scriptsize{(+2.0)}        & 76.6\scriptsize{(+8.7)}        & 74.6\scriptsize{(+2.5)}        & 75.6\scriptsize{(+6.0)}        & 74.9\scriptsize{(+5.1)}               \\
\hline
Llama-3-8B-Instruct    & 65.8        & 63.8\scriptsize{(+1.9)}        & 61.2\scriptsize{(+2.9)}        & 62.3\scriptsize{(+6.9)}        & 64.3\scriptsize{(+3.5)}        & 62.4\scriptsize{(+6.5)}        & 61.3\scriptsize{(+5.5)}        & 62.6\scriptsize{(+4.6)}               \\
\hline
Llama-3-70B-Instruct   & 73.6        & 71.2\scriptsize{(+6.4)}        & 73.0\scriptsize{(+8.6)}        & 73.2\scriptsize{(+4.2)}        & 72.1\scriptsize{(+4.9)}        & 69.5\scriptsize{(+7.7)}        & 71.0\scriptsize{(+2.2)}        & 71.7\scriptsize{(+5.7)}               \\
\hline
Llama-3.1-8B-Instruct  & 68.2        & 65.5\scriptsize{(+33.2)}        & 62.3\scriptsize{(+18.9)}        & 62.8\scriptsize{(+41.3)}        & 64.5\scriptsize{(+12.2)}        & 63.1\scriptsize{(+28.8)}        & 61.8\scriptsize{(+9.5)}        & 63.3\scriptsize{(+23.9)}               \\
\hline
Llama-3.1-70B-Instruct & 78.8        & 71.9\scriptsize{(+23.3)}        & 73.6\scriptsize{(+44.2)}        & 74.2\scriptsize{(+19.7)}        & 73.3\scriptsize{(+4.4)}        & 71.1\scriptsize{(+38.8)}        & 71.9\scriptsize{(+2.4)}        & 72.7\scriptsize{(+22.2)}    \\
        
        \bottomrule
    \end{tabular}}
    \caption{Accuracy comparison of LLMs across English and six non-English languages under default DPO and DPO with batch alignment. Improvements in non-English languages are shown in parentheses. "Rest (avg.)" is the average accuracy across non-English languages, showing enhanced cross-lingual consistency with batch alignment.}\label{tab2}
\end{table*}

\begin{table*}[t]
\centering
\small
\resizebox{\textwidth}{!}{
\begin{tabular}{lcccccc}
\toprule
\textbf{Model} & \textbf{English (95\% CI)} & \textbf{Thai (95\% CI)} & \textbf{Vietnamese (95\% CI)} & \textbf{Hungarian (95\% CI)} & \textbf{Romanian (95\% CI)} & \textbf{Rest (avg.)} \\
\midrule
Mixtral-8$\times$22B (Base) & 74.20 [72.2--76.1] & 34.80 [32.7--36.9] & 41.50 [39.4--43.7] & 42.70 [40.5--44.9] & 47.60 [45.4--49.8] & 41.7 \\
Mixtral-8$\times$22B (Ours) & 80.20 [78.4--81.9] & 37.90 [35.8--40.0] & 43.50 [41.3--45.7] & 45.90 [43.7--48.1] & 49.90 [47.7--52.1] & 44.3 \\
Llama-3.1-70B (Base)        & 73.60 [71.6--75.5] & 30.70 [28.7--32.8] & 35.00 [32.9--37.1] & 32.90 [30.9--35.0] & 36.00 [33.9--38.1] & 33.7 \\
Llama-3.1-70B (Ours)        & 79.20 [77.4--80.9] & 37.50 [35.4--39.6] & 39.10 [37.0--41.3] & 36.20 [34.1--38.3] & 38.50 [36.4--40.7] & 37.8 \\
Llama-3-70B (Base)          & 65.80 [63.7--67.8] & 28.60 [26.7--30.6] & 34.60 [32.5--36.7] & 31.50 [29.5--33.6] & 35.80 [33.7--37.9] & 32.6 \\
Llama-3-70B (Ours)          & 74.10 [72.1--76.0] & 34.10 [32.1--36.2] & 36.20 [34.1--38.3] & 36.10 [34.0--38.2] & 37.30 [35.2--39.4] & 35.9 \\
\bottomrule
\end{tabular}
}
\vspace{0.25em}

\caption{Expanded results for unseen languages. Accuracy (\%) on English and four unseen languages. Brackets show 95\% bootstrap CIs; \textit{Rest (avg.)} is the macro-average over the non-English languages.}
\label{tab:unseen-langs-ci}
\raggedright\footnotesize \textbf{Notes:} Values are accuracy (\%); brackets show 95\% bootstrap confidence intervals. Significance vs. baselines is validated with McNemar’s test.
\end{table*}

\begin{table*}[t]
    \centering
    \scalebox{0.8}{\begin{tabular}{lccccccccc}
        \toprule
        \multirow{2}{*}{\textbf{Models}} & \multicolumn{8}{c}{\textbf{ORPO Finetuning (Default)}} \\
        \cmidrule(lr){2-9}
         & \textbf{English} & \textbf{Arabic} & \textbf{Spanish} & \textbf{French} & \textbf{Japanese} & \textbf{Portuguese} & \textbf{Chinese} & \textbf{Rest (avg.)}\\
        \midrule
      Mistral-7B-Instruct-v0.1    & 60.8        & 28.3        & 55.1        & 55.3        & 45.2        & 58.6        & 54.5        & 49.5               \\
\hline
Mistral-7B-Instruct-v0.2    & 74.2        & 62.3        & 67.2        & 66.4        & 65.2        & 67.3        & 68.3        & 66.1               \\
\hline
Mistral-7B-Instruct-v0.3    & 76.6        & 63.4        & 67.3        & 68.2        & 66.4        & 72.5        & 69.6        & 67.9               \\
\hline
Mixtral-8x22B-Instruct-v0.1 & 80.2        & 64.2        & 73.5        & 74.5        & 68.3        & 73.2        & 70.5        & 70.7               \\
\hline
Llama-3-8B-Instruct    & 66.2        & 62.5        & 59.4        & 56.3        & 61.4        & 56.2        & 56.8        & 58.8               \\
\hline
Llama-3-70B-Instruct   & 74.1        & 65.8        & 65.9        & 70.2        & 68.0        & 62.3        & 68.7        & 66.8               \\
\hline
Llama-3.1-8B-Instruct  & 68.9        & 32.5        & 44.1        & 23.1        & 54.3        & 36.6        & 53.5        & 40.7               \\
\hline
Llama-3.1-70B-Instruct & 79.2        & 49.2        & 30.1        & 55.6        & 69.4        & 33.1        & 70.2        & 51.3  \\
        \midrule
        \multirow{2}{*}{\textbf{Models}} & \multicolumn{8}{c}{\textbf{ORPO + Ours (Batch Size = 7) }} \\
        \cmidrule(lr){2-9}
        & \textbf{English} & \textbf{Arabic} & \textbf{Spanish} & \textbf{French} & \textbf{Japanese} & \textbf{Portuguese} & \textbf{Chinese} & \textbf{Rest (avg.)}\\
        \midrule
     Mistral-7B-Instruct-v0.1    & 60.8        & 59.6\scriptsize{(+31.3)}        & 59.0\scriptsize{(+3.9)}        & 58.2\scriptsize{(+2.9)}        & 56.4\scriptsize{(+11.2)}        & 59.3\scriptsize{(+0.7)}        & 56.1\scriptsize{(+1.6)}        & 58.1\scriptsize{(+8.6)}               \\
\hline
Mistral-7B-Instruct-v0.2    & 74.2        & 69.1\scriptsize{(+6.8)}        & 71.1\scriptsize{(+3.9)}        & 70.9\scriptsize{(+4.5)}        & 70.1\scriptsize{(+4.9)}        & 73.3\scriptsize{(+6.0)}        & 72.3\scriptsize{(+4.0)}        & 71.1\scriptsize{(+5.0)}               \\
\hline
Mistral-7B-Instruct-v0.3    & 76.6        & 72.9\scriptsize{(+9.5)}        & 73.2\scriptsize{(+5.9)}        & 74.1\scriptsize{(+5.9)}        & 72.1\scriptsize{(+5.7)}        & 74.9\scriptsize{(+2.4)}        & 73.5\scriptsize{(+3.9)}        & 73.5\scriptsize{(+5.6)}               \\
\hline
Mixtral-8x22B-Instruct-v0.1 & 80.2        & 74.6\scriptsize{(+10.4)}        & 77.1\scriptsize{(+3.6)}        & 76.2\scriptsize{(+1.7)}        & 75.6\scriptsize{(+7.3)}        & 76.1\scriptsize{(+2.9)}        & 76.2\scriptsize{(+5.7)}        & 76.0\scriptsize{(+5.3)}               \\
\hline
Llama-3-8B-Instruct    & 66.2        & 64.2\scriptsize{(+1.7)}        & 65.6\scriptsize{(+6.2)}        & 61.2\scriptsize{(+4.9)}        & 62.3\scriptsize{(+0.9)}        & 63.5\scriptsize{(+7.3)}        & 61.2\scriptsize{(+4.4)}        & 63.0\scriptsize{(+4.2)}               \\
\hline
Llama-3-70B-Instruct   & 74.1        & 68.2\scriptsize{(+2.4)}        & 73.4\scriptsize{(+7.5)}        & 71.5\scriptsize{(+1.3)}        & 70.5\scriptsize{(+2.5)}        & 72.5\scriptsize{(+10.2)}        & 70.9\scriptsize{(+2.2)}        & 71.2\scriptsize{(+4.4)}               \\
\hline
Llama-3.1-8B-Instruct  & 68.9        & 65.1\scriptsize{(+32.6)}        & 66.5\scriptsize{(+22.4)}        & 62.3\scriptsize{(+39.2)}        & 63.4\scriptsize{(+9.1)}        & 64.5\scriptsize{(+27.9)}        & 62.9\scriptsize{(+9.4)}        & 64.1\scriptsize{(+23.4)}               \\
\hline
Llama-3.1-70B-Instruct & 79.2        & 70.2\scriptsize{(+21.0)}        & 78.2\scriptsize{(+48.1)}        & 77.6\scriptsize{(+22.0)}        & 74.9\scriptsize{(+5.5)}        & 74.2\scriptsize{(+41.1)}        & 73.5\scriptsize{(+3.3)}        & 74.8\scriptsize{(+23.5)}    \\
        
        \bottomrule
    \end{tabular}}
    \caption{Accuracy comparison of LLMs across English and six non-English languages under default ORPO and ORPO with batch alignment. Improvements in non-English languages are shown in parentheses. "Rest (avg.)" is the average accuracy across non-English languages, showing enhanced cross-lingual consistency with batch alignment.}
    \vspace{-1em}
    \label{tab3}
\end{table*}

\section{Extended Results}
\label{sec:appendix_results}

\noindent\textbf{Baseline Performance:} Table \ref{tab1} shows the performance of pre-trained LLMs (Baseline) without the application of DPO/ORPO or batch alignment method. The performance gap between English and non-English languages is significant, with an average gap of $(11.71\%)$ across all models. The largest gap is observed in the Llama-3.1-8B model, where English outperforms the average of the other languages by $(28.7\%)$. Mistral-7B models show gradual improvements across versions, with Mistral-7B-v0.3 performing better than it's previous versions, particularly in English and Portuguese. The Mixtral-8x22B model performs consistently well across languages, particularly excelling in English, Spanish, and French.

In contrast, the Llama-3 family shows varying performance, with the Llama-3-70B model outperforming the 8B version across all languages. The Llama-3.1-8B model particularly struggles with Arabic and French, while its 70B counterpart underperforms in Spanish and Portuguese, highlighting inconsistent performance of the Llama-3.1 models across different sizes and languages.

\noindent\textbf{Impact of Batch Alignment on DPO \& ORPO:}
Table \ref{tab2} compares the performance of models under default DPO settings and our proposed batch alignment method (with Batch Size = 7). Under default DPO, the performance gap between English and non-English languages increases to $(13.9\%)$, as the models become more optimized for English. However, with our batch alignment method applied to DPO, this gap is significantly reduced to an average of $(4.1\%)$, with a maximum improvement of $(23.9\%)$ observed in the Llama-3.1-8B model, which had initially struggled the most. This demonstrates that our method is particularly effective in boosting the performance of weaker models in non-English languages. Mistral-7B-Instruct v0.3 continues to show the best results in both English and non-English tasks, particularly excelling in English, Portuguese, and French, while Mixtral-8x22B also demonstrates substantial gains across all languages. Even for models that initially underperformed, such as Llama-3.1-8B, our batch alignment method leads to marked improvements in Arabic and Portuguese, highlighting the robustness of our method in addressing performance gaps across languages.

Table \ref{tab3} presents similar comparisons under the ORPO setting. With default ORPO, the performance gap between English and non-English languages increases to $(13.6\%)$, which is slightly smaller than the default DPO setting. However, with the application of our batch alignment technique, this gap is reduced to an average of $(3.6\%)$, further emphasizing the effectiveness of alignment training. Again, the Llama-3.1-8B model sees the greatest improvement, particularly in Arabic and French, where initial performance had been weakest.

\noindent\textbf{Comparing DPO and ORPO:} Tables \ref{tab2} and \ref{tab3} show that ORPO generally outperforms DPO in multilingual tasks, with an average improvement of $(1.1\%)$ for non-English languages. However, our batch alignment method significantly reduces the performance gap in both settings. The gap between English and non-English languages decreases from $(13.9\%)$ to $(4.1\%)$ in DPO, and from $(13.6\%)$ to $(3.6\%)$ in ORPO, underscoring the importance of optimizing internal reasoning consistency across languages, especially for weaker models and low-resource languages.


\begin{figure*}[!th]
  \includegraphics[width=\textwidth]{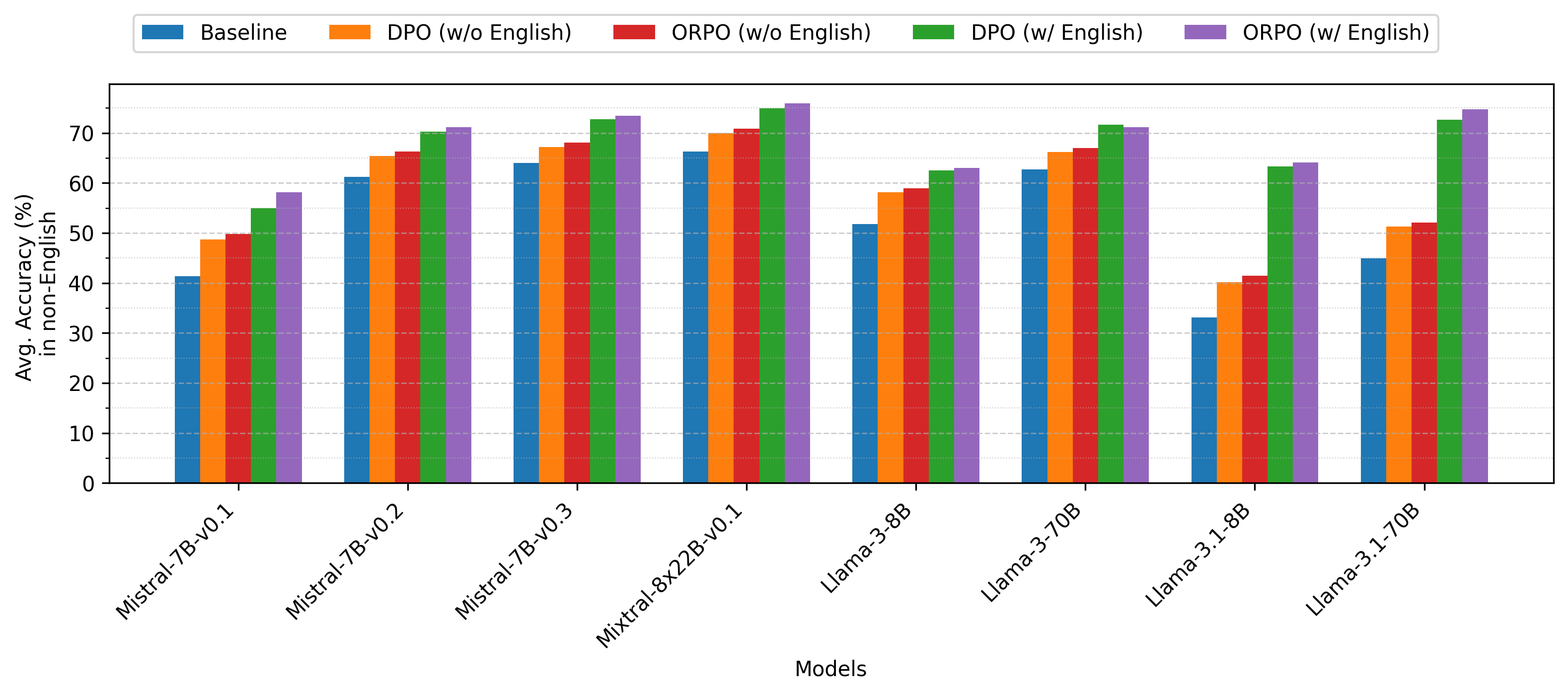}
  \caption{Impact of including vs. excluding English in training batches on non-English accuracy. Across all models, training with English (\textcolor{custompurple}{purple} and \textcolor{customgreen}{green} bars) consistently leads to higher non-English accuracy, demonstrating the role of high-resource languages in improving multilingual consistency. We use the \textit{instruct} version of each model.}
  \label{fig:w_wo_english}
\end{figure*}

\begin{table*}[!htb]
    \centering
    \scalebox{0.83}{\begin{tabular}{lcccccc}
        \toprule
        \multirow{2}{*}{\textbf{Models}} & \multicolumn{2}{c}{\textbf{Baseline}}  & \multicolumn{2}{c}{\textbf{with English (DPO)}} & \multicolumn{2}{c}{\textbf{w/o English (DPO)}}\\
        \cmidrule(lr){2-7}
         & \textbf{English} &\textbf{Rest (Avg.)} & \textbf{English} & \textbf{Rest (Avg.)} & \textbf{English} & \textbf{Rest (Avg.)} \\
        \midrule
       Mistral-7B-Instruct-v0.1 & 56.0 & 41.3 & 60.3 & 55.0 & 56.0 & 48.7 \\
        \hline
        Mistral-7B-Instruct-v0.2 & 64.0 & 61.3 & 73.5 & 70.3 & 64.0 & 65.4 \\
        \hline
         Mistral-7B-Instruct-v0.3 & 70.4 & 64.0 & 76.3 & 72.7 & 70.4 & 67.2 \\
        \hline
         Mixtral-8x22B-Instruct-v0.1 & 74.2 & 66.3 & 79.3 & 74.9 & 74.2 & 69.9 \\
        \hline
         Llama-3-8B-Instruct & 56.0 & 51.8 & 65.8 & 62.6 & 56.0 & 58.2 \\
        \hline
         Llama-3-70B-Instruct & 65.8 & 62.8 & 73.6 & 71.7 & 65.8 & 66.2 \\
        \hline
         Llama-3.1-8B-Instruct & 59.2 & 33.1 & 68.2 & 63.3 & 59.2 & 40.2 \\
        \hline
         Llama-3.1-70B-Instruct & 73.6 & 44.9 & 78.8 & 72.7 & 73.6 & 51.3 \\
        \hline
    \end{tabular}}
    \caption{Comparison of LLM's performance with and without English in the training batches under our DPO settings. "Rest (avg.)" reflects the average accuracy across non-English languages, showing that including English significantly boosts cross-lingual performance.}\label{tab6}
    \vspace{-0.5em}
  
\end{table*}

\begin{table*}[!htb]
    \centering
    \scalebox{0.83}{\begin{tabular}{lcccccc}
        \toprule
        \multirow{2}{*}{\textbf{Models}} & \multicolumn{2}{c}{\textbf{Baseline}}  & \multicolumn{2}{c}{\textbf{with English (ORPO)}} & \multicolumn{2}{c}{\textbf{w/o English (ORPO)}}\\
        \cmidrule(lr){2-7}
         & \textbf{English} &\textbf{Rest (Avg.)} & \textbf{English} & \textbf{Rest (Avg.)} & \textbf{English} & \textbf{Rest (Avg.)} \\
        \midrule
      Mistral-7B-Instruct-v0.1 & 56.0 & 41.3 & 60.8 & 58.1 & 56.0 & 49.8 \\
        \hline
         Mistral-7B-Instruct-v0.2 & 64.0 & 61.3 & 74.2 & 71.1 & 64.0 & 66.3 \\
        \hline
         Mistral-7B-Instruct-v0.3 & 70.4 & 64.0 & 76.6 & 73.5 & 70.4 & 68.1 \\
        \hline
         Mixtral-8x22B-Instruct-v0.1 & 74.2 & 66.3 & 80.2 & 76.0 & 74.2 & 70.9 \\
        \hline
         Llama-3-8B-Instruct & 56.0 & 51.8 & 66.2 & 63.0 & 56.0 & 58.9 \\
        \hline
         Llama-3-70B-Instruct & 65.8 & 62.8 & 74.1 & 71.2 & 65.8 & 66.9 \\
        \hline
         Llama-3.1-8B-Instruct & 59.2 & 33.1 & 68.9 & 64.1 & 59.2 & 41.5 \\
        \hline
         Llama-3.1-70B-Instruct & 73.6 & 44.9 & 79.2 & 74.8 & 73.6 & 52.1 \\
        \hline
    \end{tabular}}
    \caption{Comparison of LLM's performance with and without English in the training batches under our ORPO settings. "Rest (avg.)" reflects the average accuracy across non-English languages, showing that including English significantly boosts cross-lingual performance.}\label{tab7}
    \vspace{-1.2em}
\end{table*}

\paragraph{Generalization on Unseen Langauges (CI + significance).}
Table~\ref{tab:unseen-langs-ci} expands the results of RQ2 \ref{sec:rq2} (Table \ref{tab:unseen_rag}) on Thai, Vietnamese, Hungarian, and Romanian language with
\textbf{95\% bootstrap confidence intervals} ($B{=}1000$) per model and language, complementing the summary in the main text.
Gains are \textbf{consistent across all four languages} and \textbf{statistically significant} under McNemar’s test
(two-sided, $p<0.01$) for accuracy/EM. Across models, \textit{Rest (avg.)} improves by \textbf{+2.6 to +4.1} points.
For example, Mixtral-8$\times$22B improves by \textbf{+3.1} (Thai), \textbf{+2.0} (Vietnamese),
\textbf{+3.2} (Hungarian), \textbf{+2.3} (Romanian); Llama-3.1-70B improves by \textbf{+6.8}, \textbf{+4.1},
\textbf{+3.3}, \textbf{+2.5} on the same languages, while English remains stable. These results show that the
multilingual consistency gains extend to \textbf{unseen} languages and are not confined to high-resource settings.

\section{Extended Ablation Studies}
\label{sec:appendix_ablation}
We conduct a series of ablation experiments to further dissect the components of our method. Specifically, we study the impact of batch size on alignment effectiveness and investigate the use of machine-translated data as an alternative to manual translations.

\subsection{Role of High-Resource Languages} 
\label{sec:appendix_high_resource_language}

Figure \ref{fig:w_wo_english} investigates the critical role of English inclusion within training batches. Our results highlight that including English examples consistently enhances non-English accuracy by substantial margins: for example, Mixtral-8x22B improves from 66.3\% (without English) to 76.0\% (with English), and Llama-3.1-70B improves from 52.1\% to 74.8\%. This observation underscores the importance of leveraging high-resource languages like English as anchors to guide internal multilingual reasoning alignment.

As shown in Table \ref{tab6}, including English in the batch (BS = 7) significantly improved performance across most models in DPO settings. For example, the Mistral-7B-v0.1 model maintained an English score of $(56.0\%)$, but when English was excluded from the training, the average performance for other languages dropped by $(6.3\%)$, resulting in an average score of $(48.7\%)$. Similarly, the Llama-3-8B-Instruct model experienced a reduction of $(4.4\%)$, dropping to $(58.2\%)$ for non-English languages when English was not included. These results underscore the critical role of English in enhancing overall multilingual alignment and consistency.

\begin{table*}[t]
    \centering
    \scalebox{0.8}{\begin{tabular}{lcccccc}
        \toprule
        \multirow{2}{*}{\textbf{Models}} & \multicolumn{5}{c}{\textbf{Baseline (Pre-trained Performance)}} \\
        \cmidrule(lr){2-6}
         & \textbf{English} & \textbf{Thai} & \textbf{Vietnamese} & \textbf{Hungarian} & \textbf{Romanian} & \textbf{Rest (avg.)} \\
        \midrule
        Mixtral-8x22B-v0.1      & 74.2  & 34.8  & 41.5  & 42.7  & 47.6  & 41.65  \\
        \hline
        Llama-3-70B             & 65.8  & 28.6  & 34.6  & 31.5  & 35.8  & 32.63  \\
        \hline
        Llama-3.1-70B           & 73.6  & 30.7  & 35.0  & 32.9  & 36.0  & 33.65  \\
        \midrule
        \multirow{2}{*}{\textbf{Models}} & \multicolumn{6}{c}{\textbf{ORPO Finetuning (Default)}} \\
        \cmidrule(lr){2-7}
         & \textbf{English} & \textbf{Thai} & \textbf{Vietnamese} & \textbf{Hungarian} & \textbf{Romanian} & \textbf{Rest (avg.)} \\
        \midrule
        Mixtral-8x22B-Instruct-v0.1      & 80.2  & 39.5  & 45.8  & 48.2  & 52.6  & 46.53  \\
        \hline
        Llama-3-70B-Instruct             & 74.1  & 37.8  & 39.7  & 36.1  & 39.1  & 38.18  \\
        \hline
        Llama-3.1-70B-Instruct           & 79.2  & 39.6  & 43.5  & 39.9  & 42.6  & 41.40  \\
        \midrule
        \multirow{2}{*}{\textbf{Models}} & \multicolumn{6}{c}{\textbf{ORPO + Ours (Batch Size = 7)}} \\
        \cmidrule(lr){2-7}
         & \textbf{English} & \textbf{Thai} & \textbf{Vietnamese} & \textbf{Hungarian} & \textbf{Romanian} & \textbf{Rest (avg.)} \\
        \midrule
        Mixtral-8x22B-Instruct-v0.1      & 80.2  & 43.5\scriptsize{(+4.0)}  & 49.9\scriptsize{(+4.1)}  & 57.8\scriptsize{(+9.6)}  & 58.6\scriptsize{(+6.0)}  & 52.45\scriptsize{(+5.93)}  \\
        \hline
        Llama-3-70B-Instruct             & 74.1  & 41.5\scriptsize{(+3.7)}  & 45.4\scriptsize{(+5.7)}  & 41.5\scriptsize{(+5.4)}  & 46.5\scriptsize{(+7.4)}  & 43.73\scriptsize{(+5.55)}  \\
        \hline
        Llama-3.1-70B-Instruct           & 79.2  & 44.5\scriptsize{(+4.9)}  & 49.2\scriptsize{(+5.7)}  & 44.5\scriptsize{(+4.6)}  & 51.5\scriptsize{(+8.9)}  & 47.43\scriptsize{(+6.03)}  \\
        \bottomrule
    \end{tabular}}
    \caption{Accuracy comparison of LLMs across English and four non-English languages (machine-translated) under baseline, default ORPO and ORPO with batch alignment (ours). Improvements in non-English languages are shown in parentheses compared to default finetuning. "Rest (avg.)" is the average accuracy across non-English languages, showing enhanced cross-lingual consistency with batch alignment.}
    \label{tab:translated_table}
\end{table*}

In ORPO settings (Table \ref{tab7}), a similar trend was observed. Instruction-tuned models such as Mistral-7B and Llama-3.1-70B exhibited significant drops in performance when English was excluded, with reductions of $(8.3\%)$ and $(22.7\%)$, respectively. Notably, the Llama-3.1 family model showed a major decline of approximately $(22\%)$, further emphasizing the importance of high-resource languages like English for effective multilingual training. These findings highlight that including English in the training process plays a crucial role in stabilizing model performance across all languages, ensuring greater consistency and accuracy.

\begin{figure}[!tb]
  \includegraphics[width=\columnwidth]{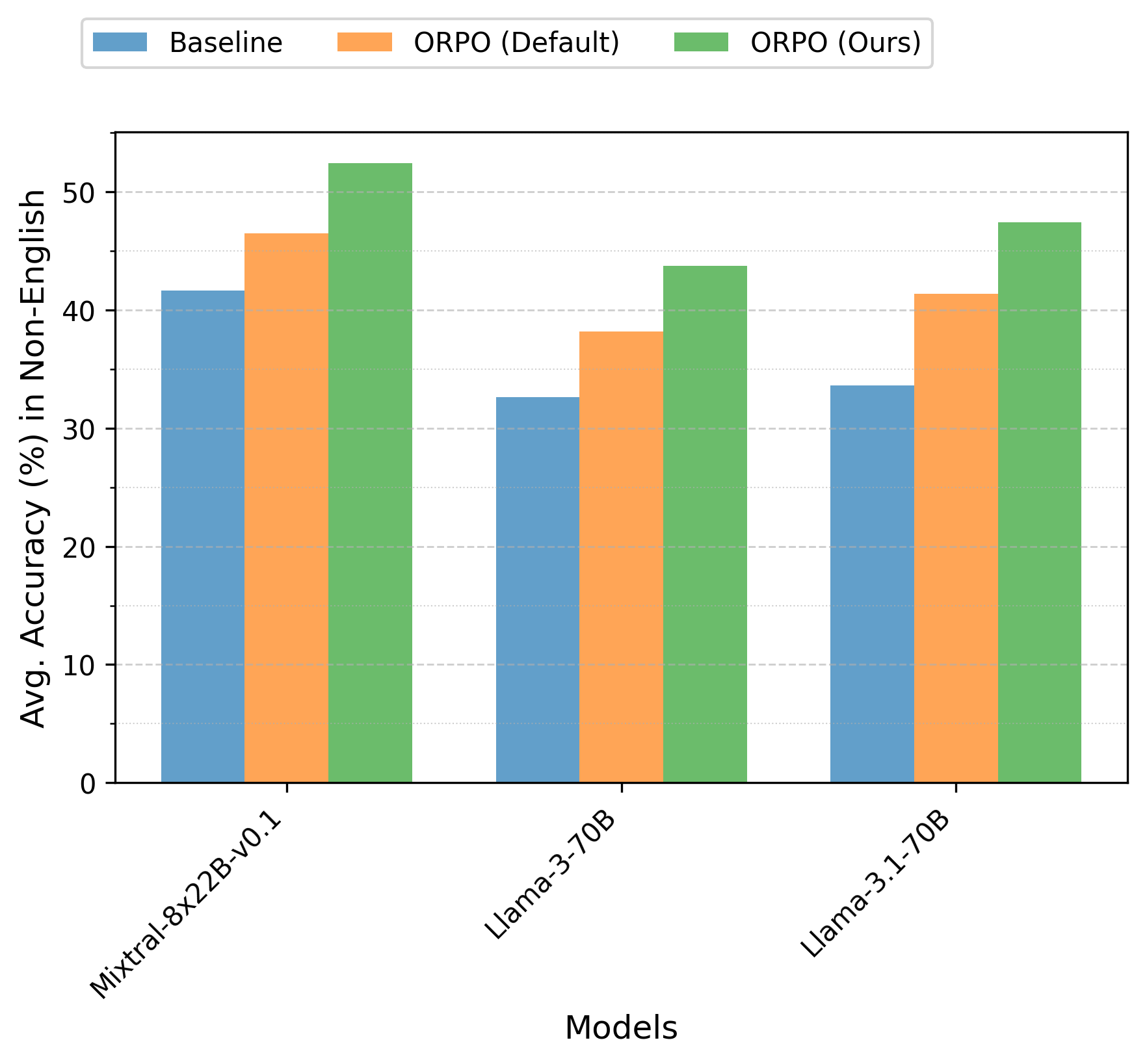}
  \caption{Illustrates avg. performance when machine translated data is used for batch-aligned finetuning of models across Thai, Vietnamese, Hungarian \& Romanian langauge.
  }
  \label{fig:machine_translated}
  \vspace{-2.5em}
\end{figure}

\subsection{Effect of Machine Translated data}
\label{sec:appendix_machine_translated}
We expand our finding \& evaluation beyond human curated datasets to machine translated data to ensure scalability. We translate the english queries and documents to Thai, Vietnamese, Hungarian \& Romanian languages, and repeat the experiments with Llama-3-70B, Llama-3.1-70B \& Mixtral-8x22B-v0.1 with ORPO only. Figure \ref{fig:machine_translated} shows the average improvement compared to the baseline \& default (ORPO). We observe consistent improvement across the three models, highlighting that machine translated data also helps to align the internal generation processess of LLM across languages, improving consistency and and making our proposed method more scalable. Though the improvement is slightly lower than the manual curated data, which could be due to the language or translation quality.

Table \ref{tab:translated_table} presents the detailed results across three models—Mixtral-8x22B-v0.1, Llama-3-70B, and Llama-3.1-70B—comparing their baseline (pre-trained performance), ORPO fine-tuning, and our proposed method for each language separately.

Our findings indicate that machine-translated data contributes to consistent improvements across all models, validating its role in scalable multilingual adaptation. Hungarian and Romanian benefit the most, with Mixtral-8x22B-v0.1 achieving a +9.6\% and +6.0\% absolute gain compared to default finetuning, while Llama-3.1-70B improves by +4.6\% and +8.9\%, respectively. This demonstrates that machine translation effectively enhances performance for morphologically complex, lower-resource languages. In contrast, Thai and Vietnamese see relatively moderate but stable gains ($\leq$6\%), potentially due to better inherent alignment with English. 

Crucially, the improvements are observed across all models, demonstrating that our proposed strategy generalizes across different architectures. The results highlight that integrating machine-translated data in pretraining and fine-tuning pipelines is a viable alternative to human-curated datasets, offering a practical solution for improving model performance across languages where high-quality parallel data is scarce. Our findings demonstrate that this approach can be directly integrated into training large-scale multilingual LLMs, ensuring cross-lingual consistency without manual intervention.

\subsection{Effect of Batch Size} 
\label{sec:appendix_batch_size}
\begin{figure}[!t]
  \includegraphics[width=\columnwidth]{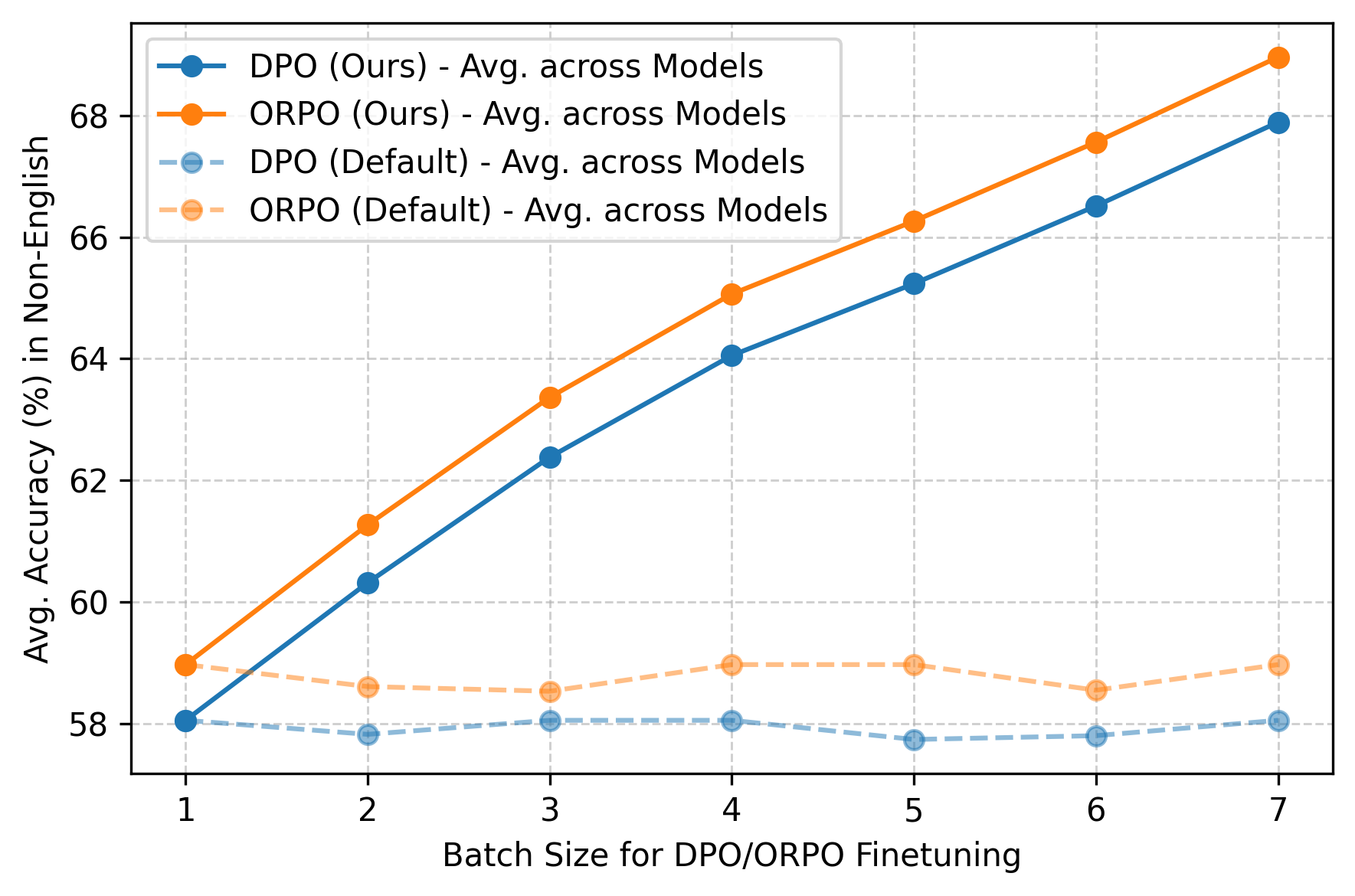}
  \caption{Illustrates how increasing batch size improves average accuracy across six non-English languages. Models with batch alignment show consistent gains, while default methods see minimal change.
  }
  \label{fig:batch_size}
  \vspace{-1.5em}
\end{figure}

Figure \ref{fig:batch_size} shows the impact of increasing batch size on average accuracy across non-English languages, with results averaged across all the models. The results highlight how our proposed batch alignment methods improve performance as batch size increases in both DPO and ORPO alignment training, whereas the default techniques show little to no effect.

\begin{table*}[!htb]
    \centering
    \scalebox{0.83}{\begin{tabular}{lcccccccc}
        \toprule
        \multirow{2}{*}{\textbf{Models}} & \multicolumn{7}{c}{\textbf{DPO Finetuning}} \\
        \cmidrule(lr){2-9}
         & \textbf{English} &\textbf{Rest (Avg.)} & \textbf{BS = 2} & \textbf{BS = 3} & \textbf{BS = 4} & \textbf{BS = 5} & \textbf{BS = 6} & \textbf{BS = 7} \\
        \midrule
        Mistral-7B-Instruct-v0.1 & 60.3 & 48.5& 50.0 & 51.4 & 52.5 & 53.2 & 54.1 & 55.0 \\ \hline
        Mistral-7B-Instruct-v0.2 & 73.5 & 65.2 & 66.4 & 67.5 & 68.3 & 68.9 & 69.6 & 70.3 \\ \hline
        Mistral-7B-Instruct-v0.3 & 76.3 & 67.0 & 68.3 & 69.5 & 70.5 & 71.2 & 71.9 & 72.7 \\ \hline
        Mixtral-8x22B-Instruct-v0.1 & 79.3 & 69.8 & 70.9 & 72.0 & 72.9 & 73.5 & 74.2 & 74.9 \\ \hline
        Llama-3-8B-Instruct & 65.8 & 58.0 & 59.1 & 60.0 & 60.8 & 61.3 & 61.9 & 62.6 \\ \hline
        Llama-3-70B-Instruct & 73.6 & 66.0 & 67.3 & 68.5 & 69.5 & 70.1 & 70.9 & 71.7 \\ \hline
        Llama-3.1-8B-Instruct & 68.2 & 39.4 & 44.9 & 49.9 & 54.0 & 56.9 & 60.0 & 63.3 \\ \hline
        Llama-3.1-70B-Instruct & 78.8 & 50.5  & 55.6 & 60.3 & 64.0 & 66.7 & 69.6 & 72.7 \\
        \midrule
      
    \end{tabular}}
    \caption{Average accuracy of LLMs in English and non-English languages as batch size increases under our DPO settings. "Rest (avg.)" indicates the average accuracy across non-English languages, showing improved cross-lingual performance with larger batch sizes. Here, BS = batch size.}
    \label{tab4}
\end{table*}

\begin{table*}[!htb]
    \centering
    \scalebox{0.83}{\begin{tabular}{lcccccccc}
        \toprule
        \multirow{2}{*}{\textbf{Models}} & \multicolumn{7}{c}{\textbf{ORPO Finetuning}} \\
        \cmidrule(lr){2-9}
         & \textbf{English} &\textbf{Rest (Avg.)} & \textbf{BS = 2} & \textbf{BS = 3} & \textbf{BS = 4} & \textbf{BS = 5} & \textbf{BS = 6} & \textbf{BS = 7} \\
        \midrule
       Mistral-7B-Instruct-v0.1 & 60.8 & 49.5 & 51.5 & 53.3 & 54.7 & 55.8 & 56.9 & 58.1 \\ \hline
        Mistral-7B-Instruct-v0.2 & 74.2& 66.1  & 67.3 & 68.3 & 69.2 & 69.8 & 70.4 & 71.1 \\ \hline
        Mistral-7B-Instruct-v0.3 & 76.6 & 67.9 & 69.2 & 70.3 & 71.3 & 72.0 & 72.7 & 73.5 \\ \hline
        Mixtral-8x22B-Instruct-v0.1 & 80.2 & 70.7 & 71.9 & 73.0 & 73.9 & 74.5 & 75.2 & 76.0 \\ \hline
       Llama-3-8B-Instruct & 66.2 & 59.7 & 58.8  & 60.6 & 61.3 & 61.9 & 62.4 & 63.0 \\ \hline
      Llama-3-70B-Instruct & 74.1 & 66.8 & 67.8 & 68.7 & 69.5 & 70.0 & 70.6 & 71.2 \\ \hline
      Llama-3.1-8B-Instruct & 68.9 &  40.7 & 46.1 & 51.0 & 55.0 & 57.8 & 60.8 & 64.1 \\ 
      \hline
      Llama-3.1-70B-Instruct & 79.2 & 51.3  & 56.7 & 61.6 & 65.6 & 68.4 & 71.5 & 74.8 \\
      \midrule
    \end{tabular}}
    \caption{Average accuracy of LLMs in English and non-English languages as batch size increases under our ORPO settings. "Rest (avg.)" indicates the average accuracy across non-English languages, showing improved cross-lingual performance with larger batch sizes. Here, BS = batch size.}\label{tab5}
    \vspace{-1em}
\end{table*}

Larger batch sizes help models capture diverse languages representing the same concept enabling more cross-lingual training signals per update, improving their ability to generalize across multilingual contexts. In contrast, default DPO and ORPO methods, which randomly select samples, show no improvement as batch size increases, plateauing at 58-60\% accuracy compared to ours which reaches upto 69\%. This is due to the lack of structured language diversity in the default settings. Our findings highlight that increasing batch diversity is crucial in cross-lingual generalization for reducing performance gaps between languages, underscoring batch diversity's role in cross-lingual generalization.

We started with English as the default (batch\_size=1), additional languages were progressively included in each batch (e.g., batch size 2 includes English and one random language, batch size 3 includes English and two random languages, and so on). Tables \ref{tab4} and \ref{tab5} summarize the impact of batch size under both DPO and ORPO settings for each model individually.

In DPO settings (Table \ref{tab4}), the Llama-3.1-8B model exhibited a notable performance increase of $(+5.5\%)$ with the inclusion of one additional language, while Llama-3-8B showed a modest gain of $(+1.1\%)$. As the batch size increased, performance improved across all models, with the Llama-3.1-8B model achieving a maximum boost of $(+23.9\%)$ at a batch size of 7. Similarly, in ORPO settings (Table \ref{tab5}), the Llama-3.1-70B model saw substantial gains of $(+23.5\%)$ at a batch size of 7, while Mixtral-8x22B-Instruct showed a rise of $(+5.3\%)$. These results indicate that larger, more diverse batches help models capture and generalize over languages with similar semantic content, particularly boosting consistent performance in non-English languages. 


\subsection{Models Size v/s Multlingual Alignment} 
\label{sec:appendix_model_size}
\begin{figure}[H]
  \includegraphics[width=\columnwidth]{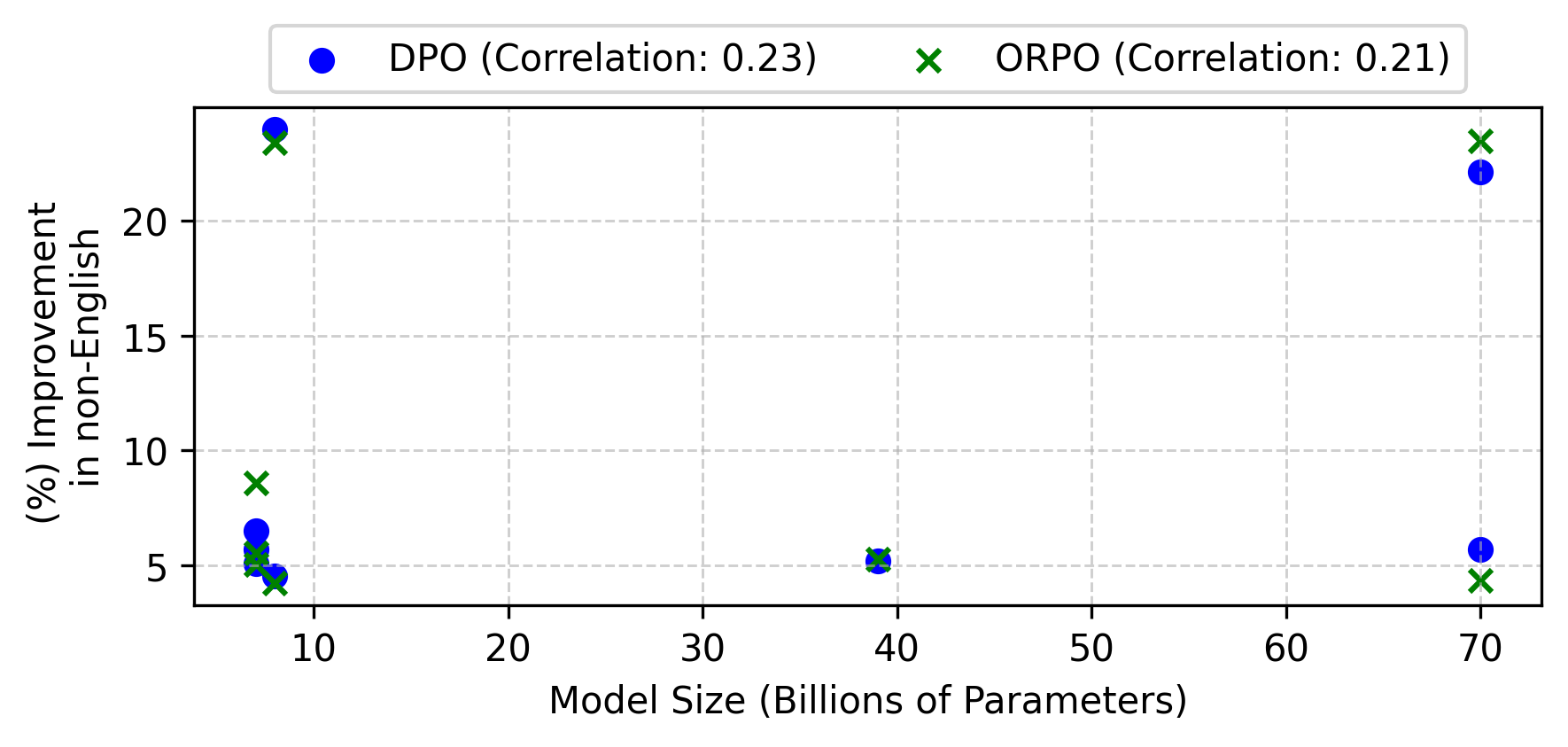}
  \caption{Shows the relationship between model size and average performance improvement in non-English languages under DPO and ORPO settings. The modest correlation observed demonstrates that factors beyond model size contribute to performance gains.}
  \label{fig:corelation}
\end{figure}

Figure \ref{fig:corelation} shows the relationship between model size and language improvement for both DPO and ORPO under our alignment strategy. While both DPO \& ORPO shows a positive corelation of 0.23 and 0.21 respectively, it is not particularly strong, demonstrating that larger models do tend to improve language performance, but the relationship isn't very pronounced. This could indicate that simply increasing model size may not completely close the language gap, and other factors (such as specific fine-tuning strategies) may play a role.  

\end{document}